\documentclass{vgtc}                 % review
\ifpdf%                                % if we use pdflatex
  \pdfoutput=1\relax                   % create PDFs from pdfLaTeX
  \usepackage{graphicx}                % allow us to embed graphics files
  \DeclareGraphicsExtensions{.pdf,.png,.jpg,.jpeg} % for pdflatex we expect .pdf, .png, or .jpg files
\else%                                 % else we use pure latex
  \usepackage{graphicx}                % allow us to embed graphics files
  \DeclareGraphicsExtensions{.eps}     % for pure latex we expect eps files
\fi%

\graphicspath{{figures/}{pictures/}{images/}{./}} % where to search for the images
\usepackage{microtype}                 % use micro-typography (slightly more compact, better to read)
\PassOptionsToPackage{warn}{textcomp}  % to address font issues with \textrightarrow
\usepackage{textcomp}                  % use better special symbols
\usepackage{mathptmx}                  % use matching math font
\usepackage{times}                     % we use Times as the main font
\usepackage{cite}                      % needed to automatically sort the references
\usepackage{tabu}                      % only used for the table example
\usepackage{booktabs}                  % only used for the table example
\usepackage{epsfig}
\usepackage{tabularx}
\usepackage{amsmath}
\usepackage{amssymb}
\usepackage{color}
\usepackage{dblfloatfix}
\usepackage{fancyhdr}

\onlineid{1183}

\vgtccategory{Research}

\vgtcinsertpkg

\title{Collaborative Augmented Reality on Smartphones\\via Life-long City-scale Maps}

\author{Lukas Platinsky %
\and Michal Szabados % \thanks{e-mail: mszabados@lyft.com}\\ %
\and Filip Hlasek % \thanks{e-mail: fhlasek@lyft.com}\\ %
\and Ross Hemsley % \thanks{e-mail: rhemsley@lyft.com}\\ %
\and Luca Del Pero % \thanks{e-mail: ldelpero@lyft.com}\\ %
\and Andrej Pancik % \thanks{e-mail: andrej@bluevisionlabs.com}\\ %
\and Bryan Baum % \thanks{e-mail: bryan@bluevisionlabs.com}\\ %
\and Hugo Grimmett % \thanks{e-mail: hgrimmett@lyft.com}\\ %
\and Peter Ondruska\thanks{This work was conducted at Blue Vision Labs (now part of Lyft Level 5). To contact authors please email: peter@ondruska.com} %
}

\teaser{
    \centering
    \includegraphics[width=53mm]{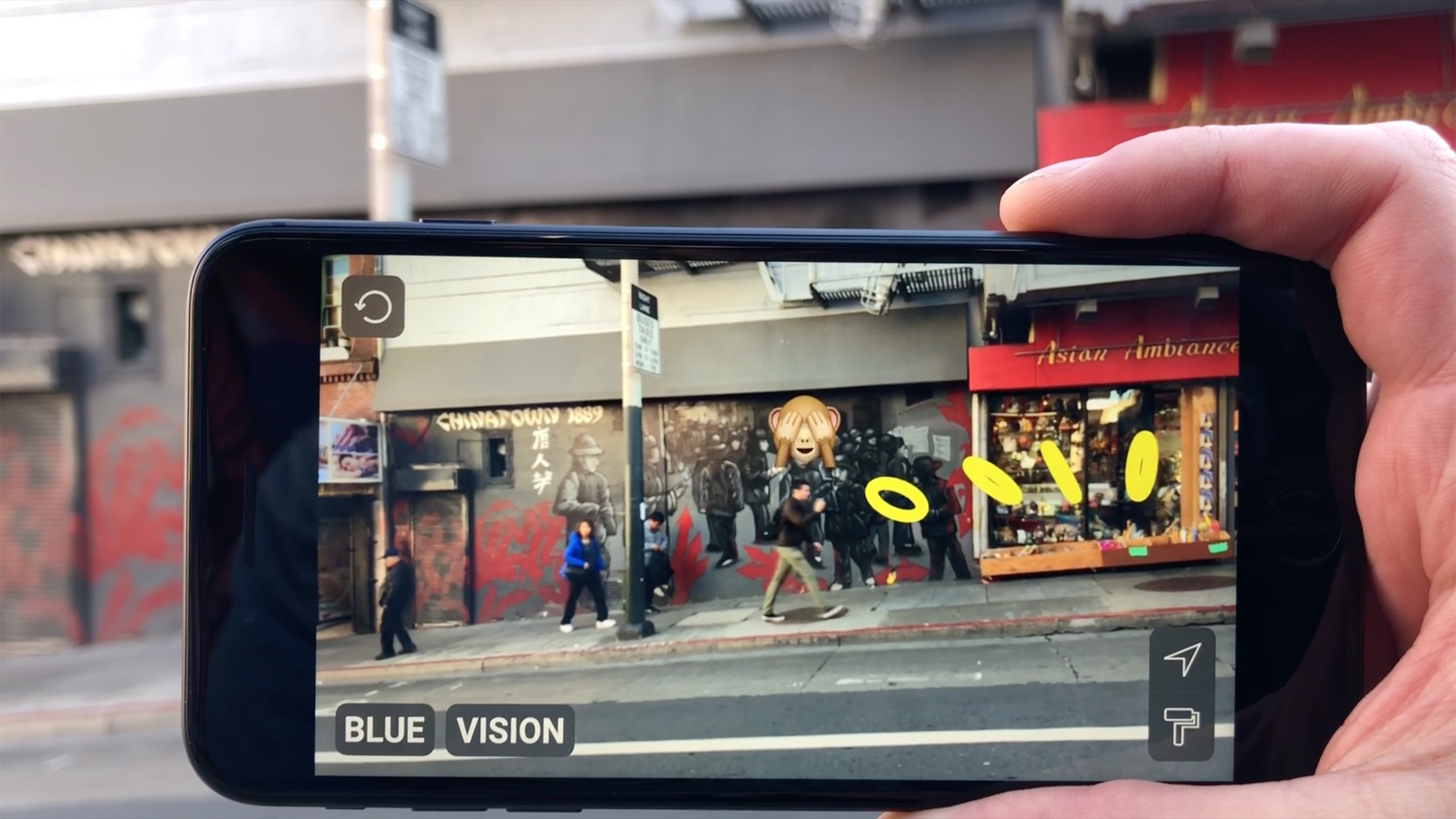}
    \includegraphics[width=53mm]{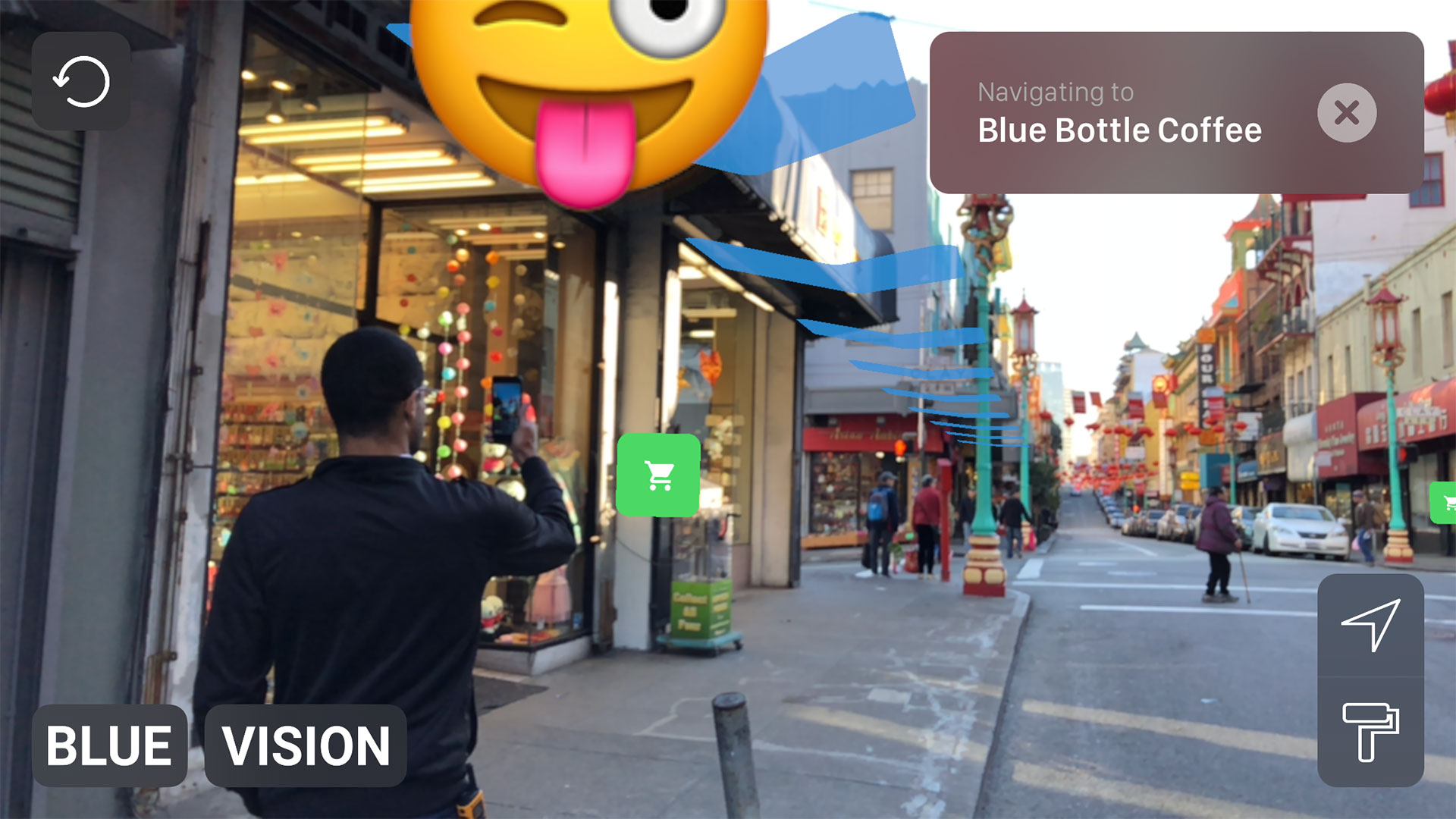}
    \includegraphics[width=53mm]{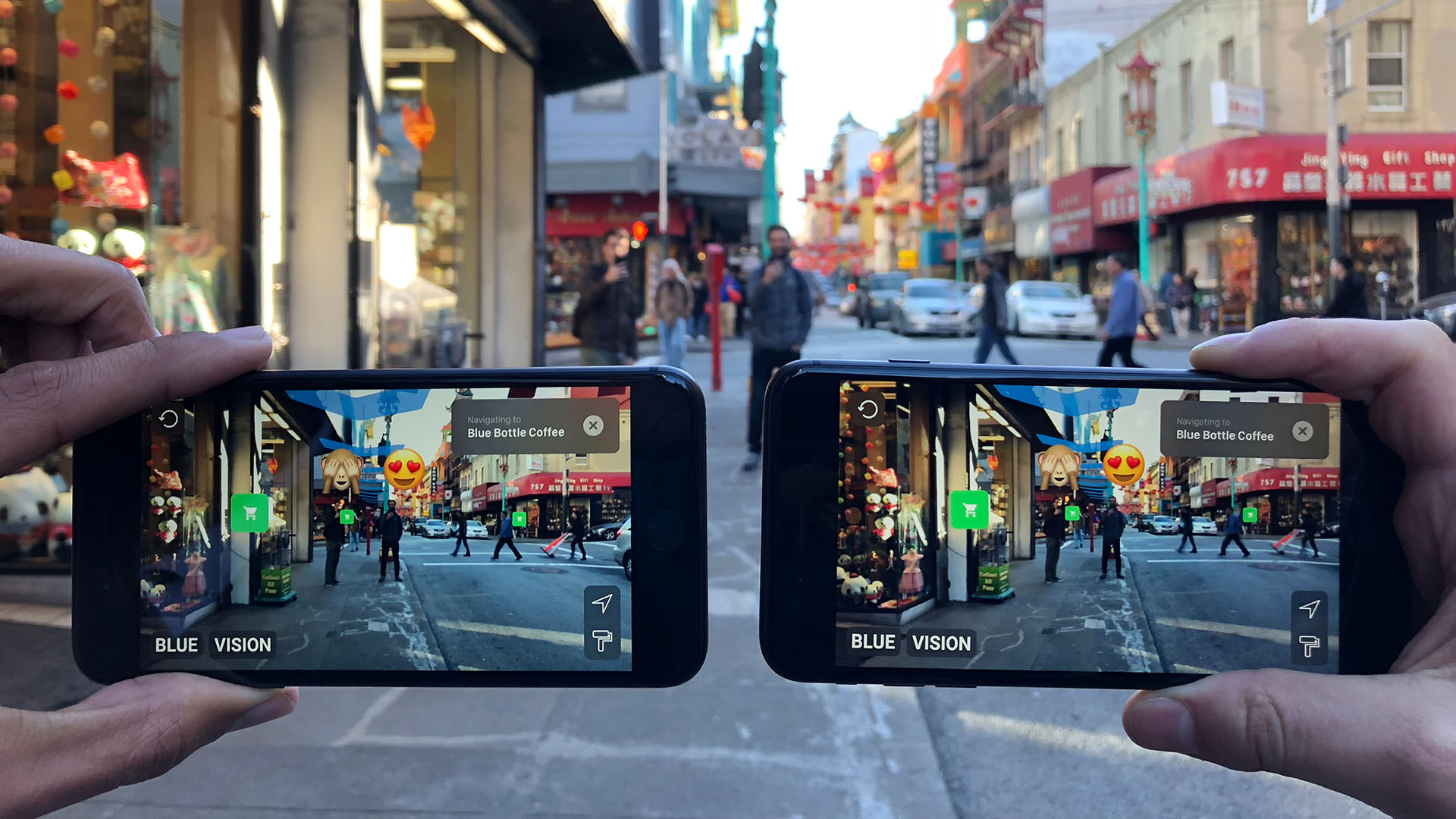}
    \includegraphics[width=53mm]{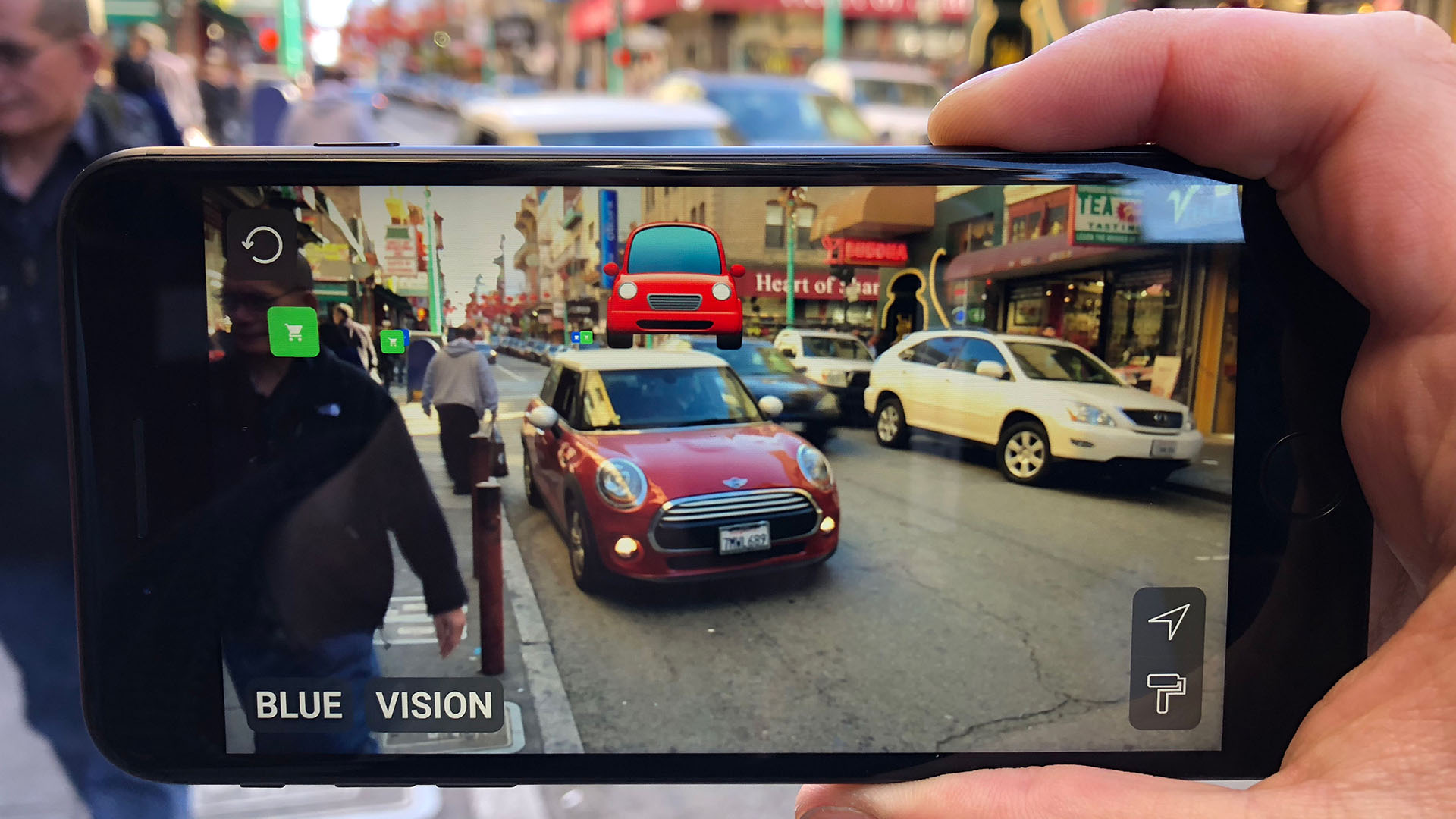}
    \includegraphics[width=53mm]{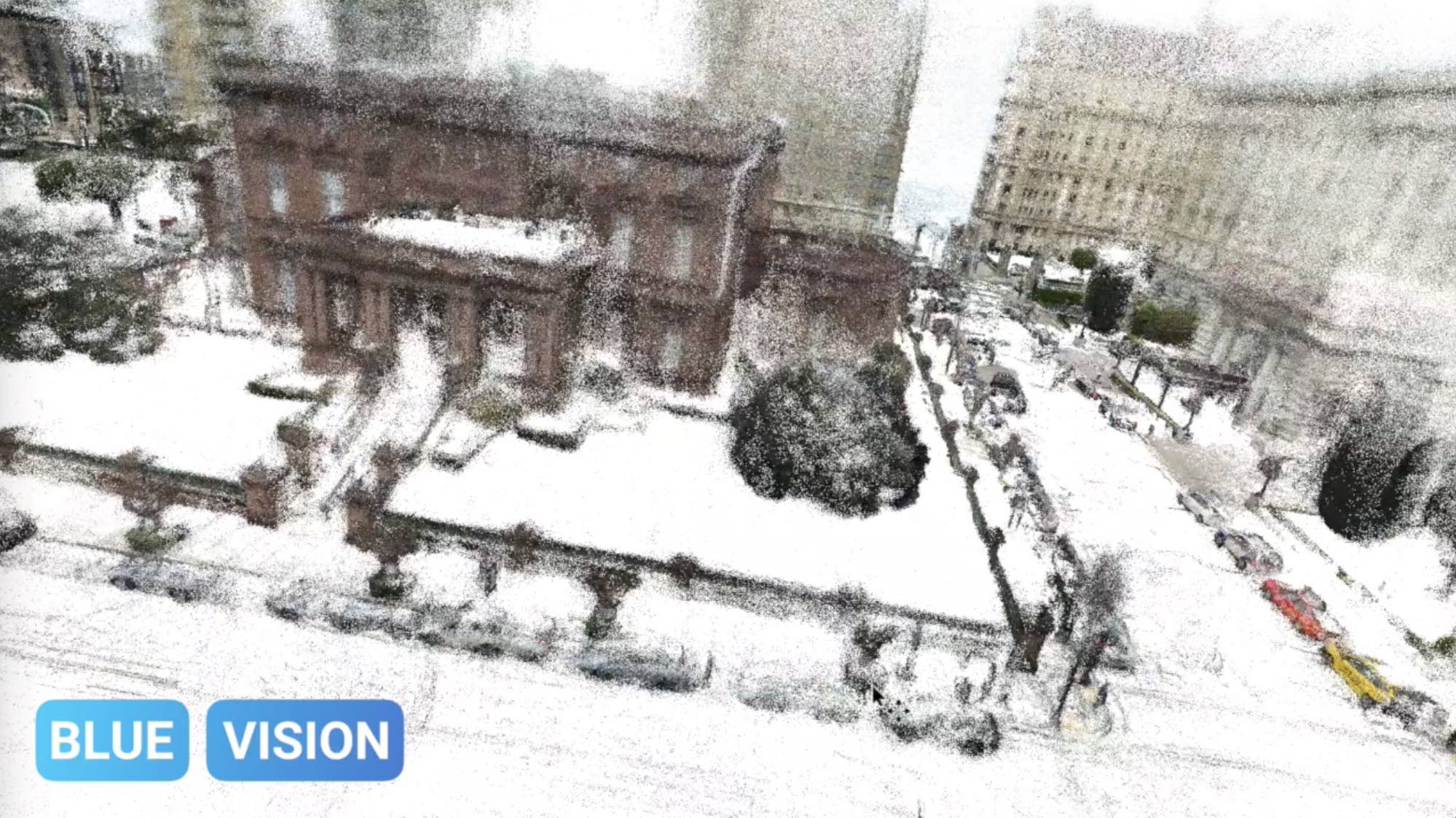}
    \includegraphics[width=53mm]{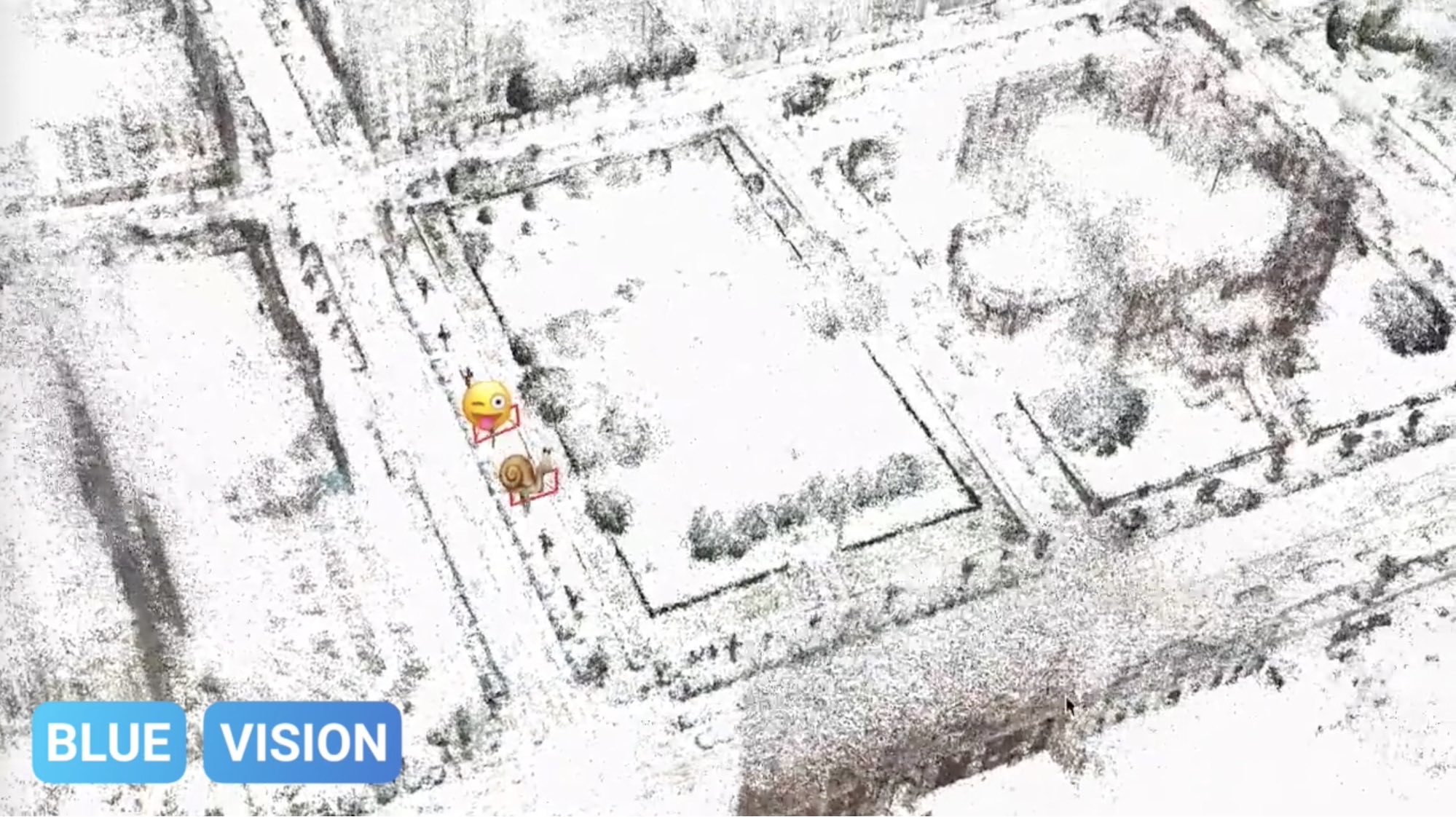}
    \caption{Shared and collaborative augmented reality experience on mobile devices, powered by high-precision localisation in city-scale 3D maps.}
    \label{fig:intro}
}

\abstract{In this paper we present the first published end-to-end production computer-vision system for powering city-scale shared augmented reality experiences on mobile devices. In doing so we propose a new formulation for an experience-based mapping framework as an effective solution to the key issues of city-scale SLAM scalability, robustness, map updates and all-time all-weather performance required by a production system. Furthermore, we propose an effective way of synchronising SLAM systems to deliver seamless real-time localisation of multiple edge devices at the same time. All this in the presence of network latency and bandwidth limitations. The resulting system is deployed and tested at scale in San Francisco where it delivers AR experiences in a mapped area of several hundred kilometers.
To foster further development of this area we offer the data set to the public, constituting the largest of this kind to date.
}

\CCScatlist{
  \CCScatTwelve{Computing methodologies}{Artificial intelligence}{Computer vision}{Tracking and Reconstruction};
  \CCScatTwelve{Computing methodologies}{Mixed / augmented Reality}{}{}
}


\begin{document}
\pagestyle{fancy}

\maketitle
\thispagestyle{fancy}

\begin{figure*}[b]
\centering
\includegraphics[width=175mm]{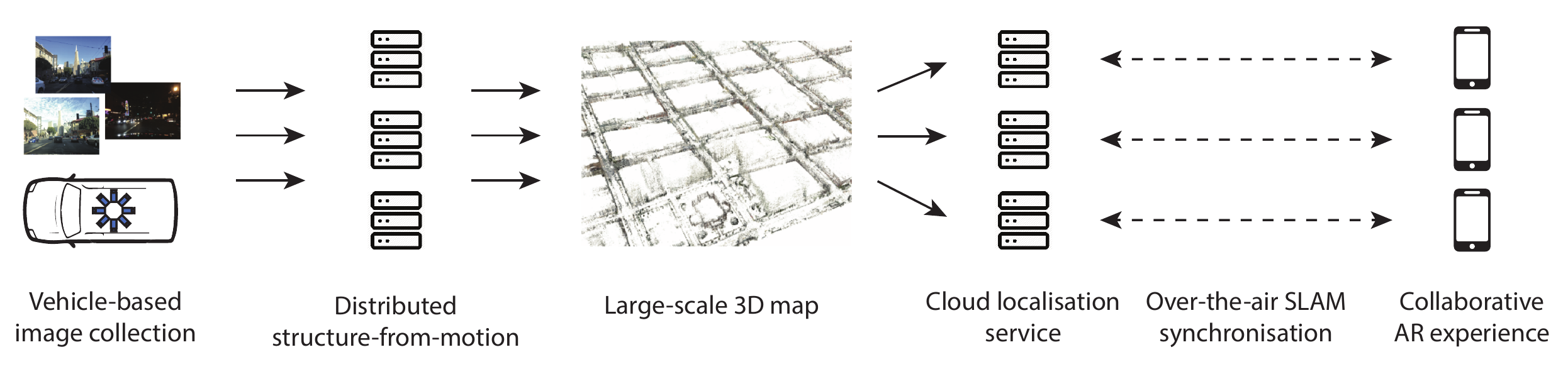}
\caption{The system diagram for the proposed system. A large-scale 3D map is computed from a vehicle-based image collection and used to localise camera-equipped edge devices for AR experiences over a mobile network.}
\label{fig:system}
\end{figure*}

%%%%%%%%% BODY TEXT
\section{Introduction}

Most of the currently available production Augmented Reality (AR) today are  single-user experiences running at a small scale. For example, in games such as Pokemon Go, a user can see AR content on their edge device (e.g. smartphone or AR glasses), but a second user cannot see the same content in the same place on their own device. This fundamental limitation is due to the each device's pose being estimated in its own coordinate systems different to each other, by the simultaneous localisation and mapping (SLAM) system running on-board. Additionally, these are often limited to small scale (i.e. one room), constant visual conditions, and do not support large-scale applications such as outdoor AR navigation.

The properties of SLAM systems are well known \cite{strasdat2012visual, fuentes2015visual, taketomi2017visual}, but building production-grade, city-size, and multi-agent solutions poses unique challenges due to the high requirements for system robustness, scalability, multi-weather and time-of-the day performance, and map updates combination of which was never explored together.

In this work we present the first published end-to-end production system that allows city-scale robust localisation of multiple devices in the same shared coordinate system for building shared and collaborative augmented reality experiences (as shown in Figure \ref{fig:intro}, and the accompanying video \footnote{Video of the deployed system is at \url{https://youtu.be/tXwVg2S9wuY}.}). In doing so we propose and exploit a new experience-based mapping framework for constructing large-scale all-time all-weather maps, keeping them perpetually updated and supporting localisation of edge devices over a mobile network. In particular, the contributions of this paper are five-fold:
\begin{enumerate}
\itemsep-0.2em 
\item The end-to-end design of a production-grade system for augmented reality at scale.
\item An experience-based mapping framework, which is an effective solution to the key issues of all-time all-weather city-scale SLAM scalability.
\item A novel way of localising multiple edge devices in the same coordinate space, in real time over a mobile network with latency and limited bandwidth.
\item A comprehensive evaluation of map-building and AR localisation performance on a city-scale dataset.
\item A new city-scale dataset \footnote{To request access to the dataset please visit \url{www.bluevisionlabs.org}}.
\end{enumerate}

\section{Related work}
% In this paper we present a large-scale all-times all-weather visual-localisation-based system for building AR experiences.

Some previous work tackles global localisation without using a map of the environment, for example by combining GPS and inertial sensors (e.g. \cite{5336489}), or GPS and visual odometry (e.g., in \cite{6738516}). However, in our tests neither of these options met the accuracy requirements for providing immersive real-time AR experiences due to large GPS location error, particularly in urban settings with tall buildings (urban canyons). To obtain the required accuracy, we have used a visual mapping and localisation approach providing a more accurate localisation signal than GPS. This approach requires constructing a large-scale global map of the environment before the AR experiences can happen.

Much work has been done on building mapping and localisation algorithms. Many of the classical systems \cite{davison2007monoslam, murTRO2015, engel2014lsd} have been achieving impressive results with single-agent real-time systems performing mapping and localisation at the same time (SLAM). There are also several impressive solutions that specifically provide the ability to build a map first and then use it to localise allowing also for multi-agent solutions \cite{lynen2015get, schneider2018maplab, forsterCollabSLAM}. These systems, however, typically operate in a limited geographic area due to the memory constraints and algorithmic complexity of the underlying structure-from-motion (SfM) problem. Their map-building part is often limited to a single machine and cannot be directly used with the amounts of data processed by the solution proposed in this paper.

Building visual maps at a scale needed for the described use cases has been addressed in \cite{Agarwal2011BRD, klingner2013street, Zhu2017ParallelSF}. Compared to these approaches, our method is designed to optimally exploit the multi-experience nature of the data collection leading to approximately linear complexity of map-building process, fast map updates, and extreme robustness to outliers and SfM failures arising from difficult visual conditions. Unlike the previous solutions, the proposed system is significantly more robust to the variances in quality of the input data, allowing for crowd-sourced approach for data collection. Moreover, we do not stop at the map-building stage and provide a large-scale multi-agent localization framework as well. 

Image-based location retrieval is a topic with a lot of interest and it was explored, for example, in \cite{irschara2009structure, sattler2011fast, sattler2012improving, svarm2016city, feng2015fast, cummins2010fab, kendall2015posenet, li2012worldwide, liu2017efficient, Zeisl_2015_ICCV}. Many of the approaches focus on limited geographic areas or on a limited set of lighting conditions. Several impressive works focusing on providing localisation of mobile devices presented in \cite{monoMobileSLAM, scalable6DOF, wideTAM} suffer from similar vices, allowing for experiences only in significantly limited areas and we show several orders of magnitude improvement in the size of the area supported in our experiments. We present a multi-agent system where many devices continuously interact and localise within a single, unbounded map under varying visual conditions.

Vision systems degrade in changing visual conditions due to increasing perceptual difference (i.e. day vs night) between the map and localisation images. Robust, all time-time, all-weather performance is key for production systems and was considered in \cite{linegar2016made, arandjelovic2016netvlad, sattler2018benchmarking} by building robust localisation feature representations. In contrast, we take inspiration from experience-based navigation framework \cite{ChurchillIJRR2013} where robust multi-weather performance is achieved by repeatedly collecting data along a route in varied conditions and building one multi-weather map. By doing so the map covers all possible visual conditions resulting into robust performance. This setup is uniquely beneficial for building productions systems, where more data can be collected as required to achieve the desired performance of the system. In particular, in this work we focus on the ability to combine these experiences, collected at scale by a fleet of vehicles, robustly fuse them to form a single map, and efficiently update the map when more data are available.

Finally, unlike all previous works that addressed only parts of the technical problem, this is the first work that ties all the components together and at scale into one production-grade AR system. We acknowledge imperfections in the current algorithms and our collected data, and rather than attempting to find a silver bullet for each part of the system, we acknowledge the imperfections and build a redundant and robust solution around it.

We are also releasing a new large-scale mapping and localisation dataset from San Francisco, comprising millions of frames and videos. It is the largest publicly-available dataset of this kind, 5x larger than the next largest similar dataset \cite{ChenCVPR2011}.

Main difference - many orders of magnitude more. Focus is not on solving each individual part perfectly. We accept imperfect nature of algorithms and design the whole end-to-end approach with redundancy and solution focused on manipulating several orders of magnitude larger data than the previous known approaches. 

%\begin{figure*}
%    \centering
%    \includegraphics[width=180mm]{pipeline}
%    \caption{The map-building pipeline. First, the data are split into highly-redundant batches that are processed in parallel and assembled into a single map. Thanks to this parallelism and redundancy, the system is highly robust to various SfM failures and linearly scalable.}
%    \label{fig:mapbuilding}
%\end{figure*}

\section{System overview}
\label{sec:mapbuildsystem}
We propose a computer vision system that enables collaborative augmented reality experiences. At its core lies a map-based localisation system outlined in Figure \ref{fig:system} that can track the global pose of an edge device (e.g. smartphone or AR glasses) $p^\mathcal{M}_t$ over time in the shared coordinate space of the 3D map $\mathcal{M}$ of the environment. This leverages the approximate location $p^{gps}_t$ given by the device GPS, and the image history of what it sees through its camera $I_{1 \dots t}$.

The map is constructed from a large set of individual data collection logs called \emph{experiences} $\{\mathcal{I}^1 \dots \mathcal{I}^N \}$. Each experience comprises a set of images $\mathcal{I}^i = \{I^i_1 \dots I^i_O \}$ and GPS points collected during a single session at particular time of day / weather condition by a vehicle equipped with multiple cameras and an INS system. It can range in length from several meters to up to a few kilometers. The ability to build maps from very large number of experiences is the key to achieving high performance and robustness (see Sec.~\ref{sec:mapping}). We use a fleet of cars to collect experiences that are used in map-building to efficiently control and improve the performance of the resulting system.

\begin{figure*}[t]
    \centering
    \includegraphics[width=175mm]{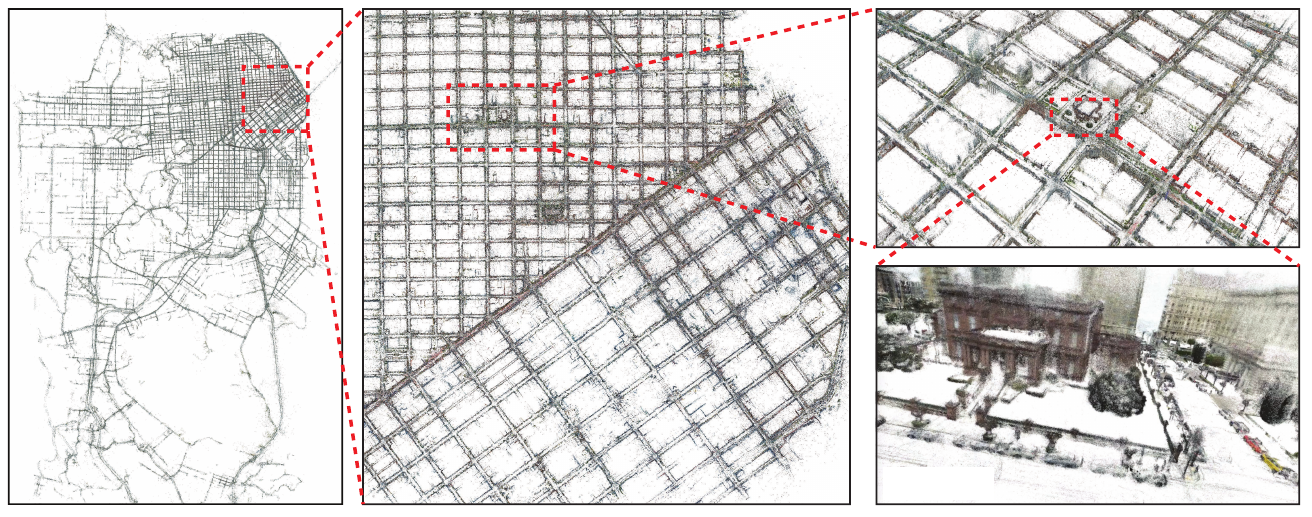}
    \caption{The city-scale map built for supporting AR experiences. The subset of the map we evaluated AR performance (middle) consists of 4249 submaps and was computed on a cluster of 500 CPUs over period of 91 hours.}
    \label{fig:map}
\end{figure*}

The map contains billions of landmarks, and due to its size cannot be stored on the edge device. Instead, the device runs a visual-inertial odometry (VIO) system locally. This system is aligned with the map through the local-to-global coordinate transform $T_{vo \rightarrow \mathcal{M}}$, recovered by estimating the pose with respect to the map of a subset of the individual images captured by the device. Due to map size and the need to keep it perpetual updates, this takes place in the cloud. In Section \ref{sec:localisation} we describe how we dealt with network latency and limited network bandwidth.

This localisation system is then augmented with \textit{persistent content storage} and \textit{real-time pose sharing}, enabling powerful collaborative AR experiences. We describe details of these in Section \ref{sec:experiments}. In what follows we present the core architecture of the system, while we discuss smaller implementations details in \ref{sec:experiments}.

\section{Large scale experience-based mapping}
\label{sec:mapping}

In this section we present our system for building a large scale map $\mathcal{M}$ from the set of collected experiences $\mathcal{I}_1 \dots \mathcal{I}_N$. It is designed to achieve three key goals:
\begin{enumerate}
    \itemsep-0.1em 
    \item \emph{Localisation performance}: reliable all-time, all-weather localisation
    \item \emph{Map robustness}: ensure map quality over a large geographic area in the face of SfM failures
    \item \emph{Map freshness}: efficient map updates by adding more data when needed.
\end{enumerate}

The algorithm consists of four components: (1) data splitting, (2) submap building, (3) submap verification, and (4) submap fusion with map update. We now provide an overview before presenting each component in detail.

The algorithm starts with an empty map $\mathcal{M}$ and then iteratively adds new experiences to it. For each experience, we group all its images into smaller overlapping subsets (data splitting), and then run SfM on each subset independently, producing submaps (submap building). This is the most expensive part of the algorithm, but we parallelise it on a cluster of CPUs. Therefore, there is no limit to the size of input experiences. Next, SfM failures are identified, and these `broken' submaps are discarded. The remaining submaps are fused into a single, global map. The entire map can be built from scratch in approximately linear time complexity with the amount of data, or alternatively, experiences can be added one-by-one. The resulting map $\mathcal{M}$ comprises $K$ geo-aligned submaps $\mathcal{M}_{1 \dots K}$, each containing the camera poses $p$ of the input images, and a set of 3D localisation landmarks $x$ such that $\mathcal{M}_i = \{ p_{1 \dots M}, x_{1 \dots L} \}$.

\begin{figure}[bh]
    \centering
    \includegraphics[width=75mm]{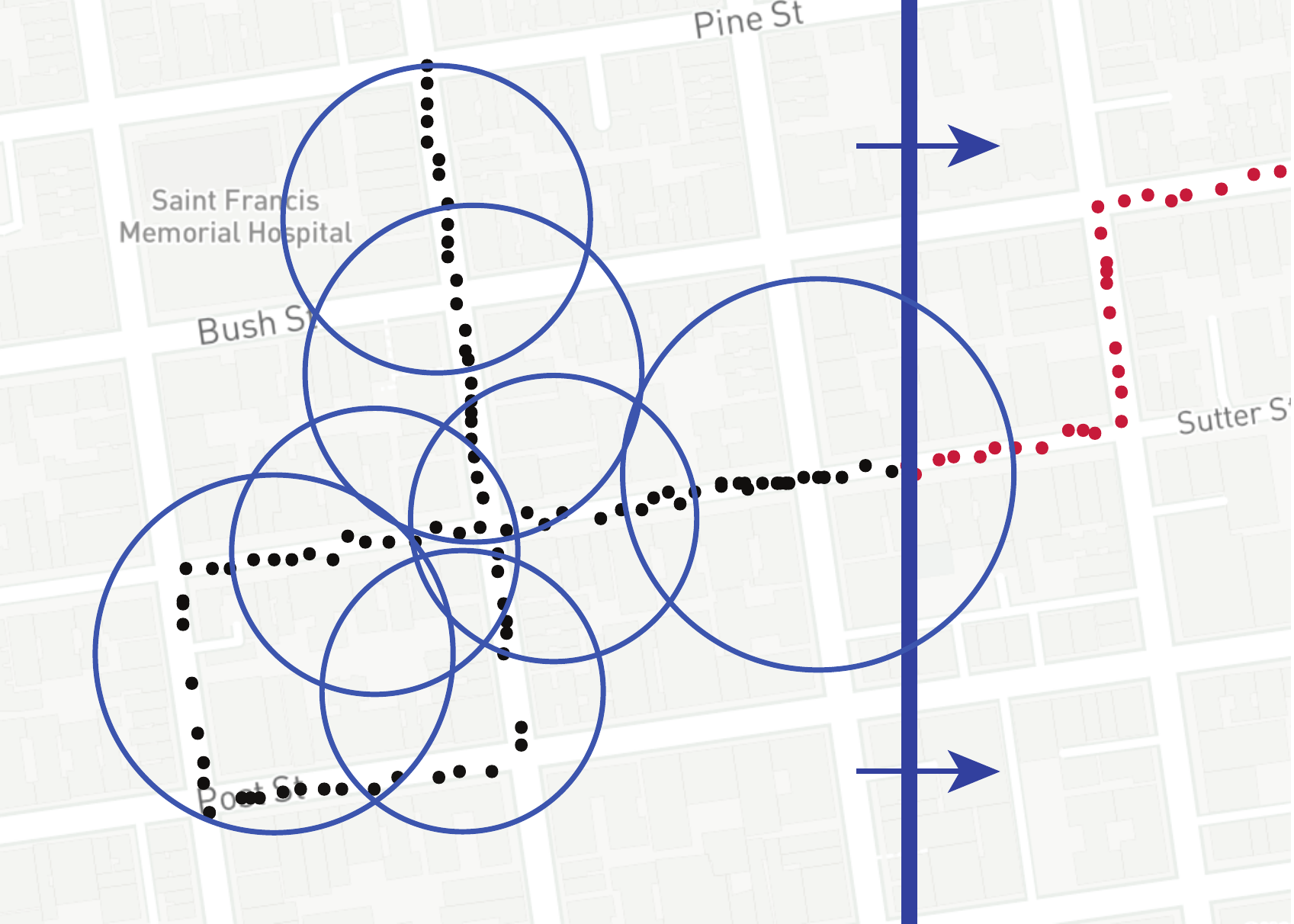}
    \caption{The line-sweep algorithm that splits data collected during one mapping session into submaps of size up to 1,000 frames.}
    \label{fig:splitting}
\end{figure}

\textbf{1. Data splitting.}
We bound the computational complexity of submap building (step 2) by grouping the images in an experience (which can be potentially numerous) into consistently-sized subsets of images. The first image in the experience is assigned to a subset, which is grown in size until it contains 1,000 images (we enforce subsets to have a circular shape, with a minimum radius of 20m). This is done using a greedy line-sweep algorithm over the 2D GPS positions of each image (see Figure \ref{fig:splitting}). In areas with large image density, we randomly sub-sample images, to fit into the 1,000 limit. These values were chosen empirically to allow submaps to cover enough geographical area (making the grouping robust to GPS errors, which are severe in urban canyons), yet be small enough for efficient submap building. Lastly, for each submap in an experience we add images randomly chosen from neighbouring submaps, which is key for submap fusion (step 3). To enable the fusion of submaps across experiences, we also add randomly-selected images from the same geographic location but different experiences. While this increases the computation required in step 2 (most images are in multiple submaps), this greatly simplifies the fusion step.

\begin{figure}[b]
    \centering
    \includegraphics[width=80mm]{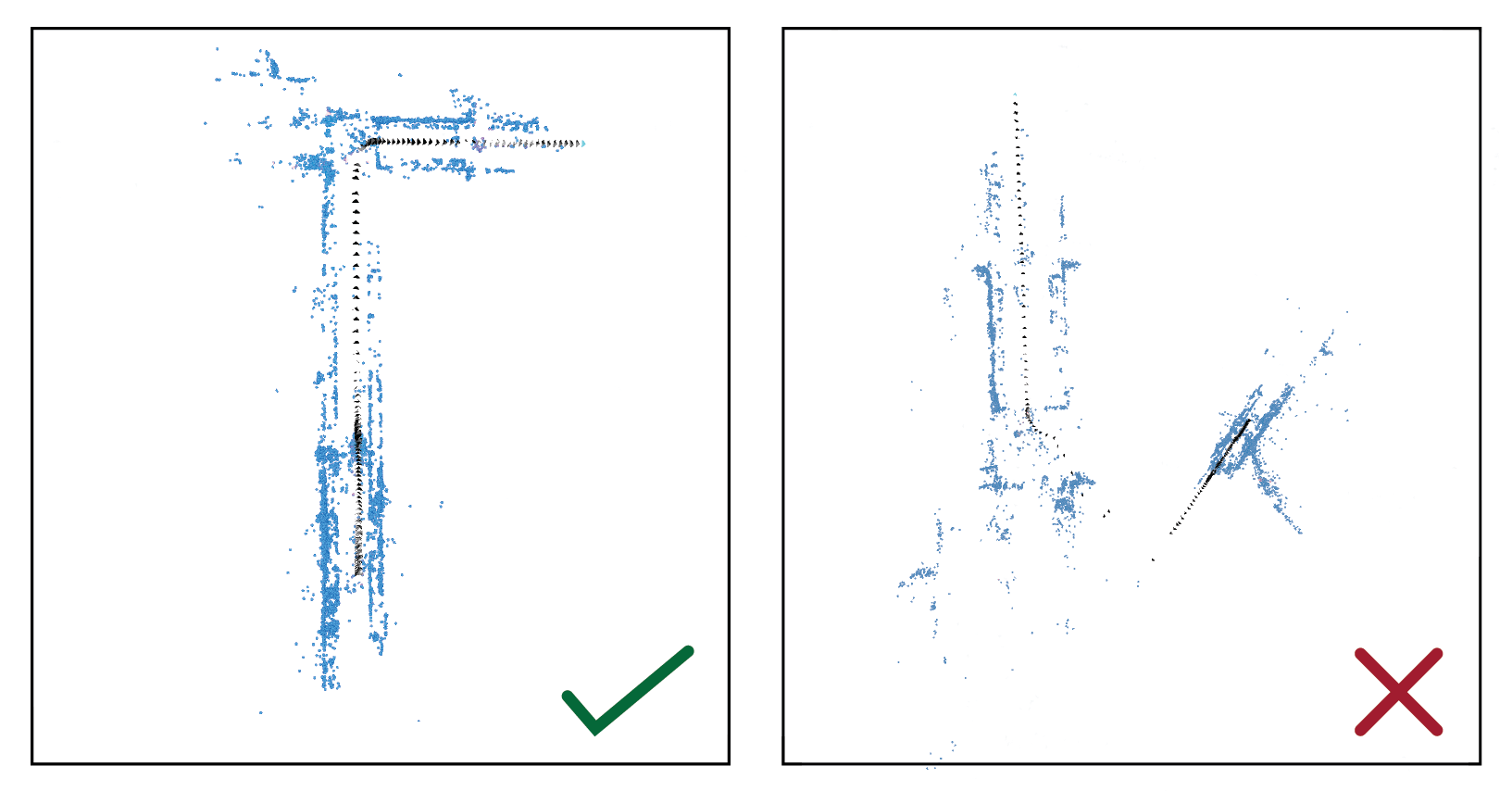}
    \caption{Examples of submaps the map is composed of. On the \emph{left} a submap where SfM succeeded, and on the \emph{right} one where structure-from-motion failed due to difficult visual conditions. Failed submaps are discarded prior to submap fusion. }
    \label{fig:submaps}
\end{figure}

\textbf{2. Submap building.}
Next, incremental SfM \cite{moulon2012adaptive} is run on each subset independently, relying on a standard pipeline of feature extraction, matching, track creation and bundle adjustment. This recovers the poses of the images and the 3D positions of landmarks, by minimising the re-projection error:
\begin{equation}
    \mathcal{C}(x,p) = \sum_{i,j} (z_{i,j}-\pi(x_i,p_j)) ^ 2 + \lambda \sum_i (p_i - p_i^{gps})^2
\label{eq:reprojection}
\end{equation}
where $z_{i,j}$ is the position of the $i$-th feature in the $j$-th image, and $\pi(x_i,p_j)$ is re-projection of landmark $x_i$ using pose $p_j$. The last term is a GPS prior, weighted by $\lambda$. Thanks to the small size of the subsets, each submap can be built on a single CPU. Examples of submaps are shown in Figure \ref{fig:submaps}.

\textbf{3. Submap verification.}
Due to challenging conditions such as occlusions, moving objects, or camera failures, SfM can output incorrect or warped 3D maps as shown in Figure \ref{fig:submaps}. To detect this we check whether the reconstructed poses agree with the relative orientations of consecutive images as measured by the INS. Submaps are also discarded if the reconstructed gravity vectors deviate too much from the INS deviate measurement, or if the positions of consecutive images violate the kinematic constraints of a simple vehicle motion model. Discarded submaps are not used for map fusion or update.

\begin{figure}[t]
    \centering
    \includegraphics[width=80mm]{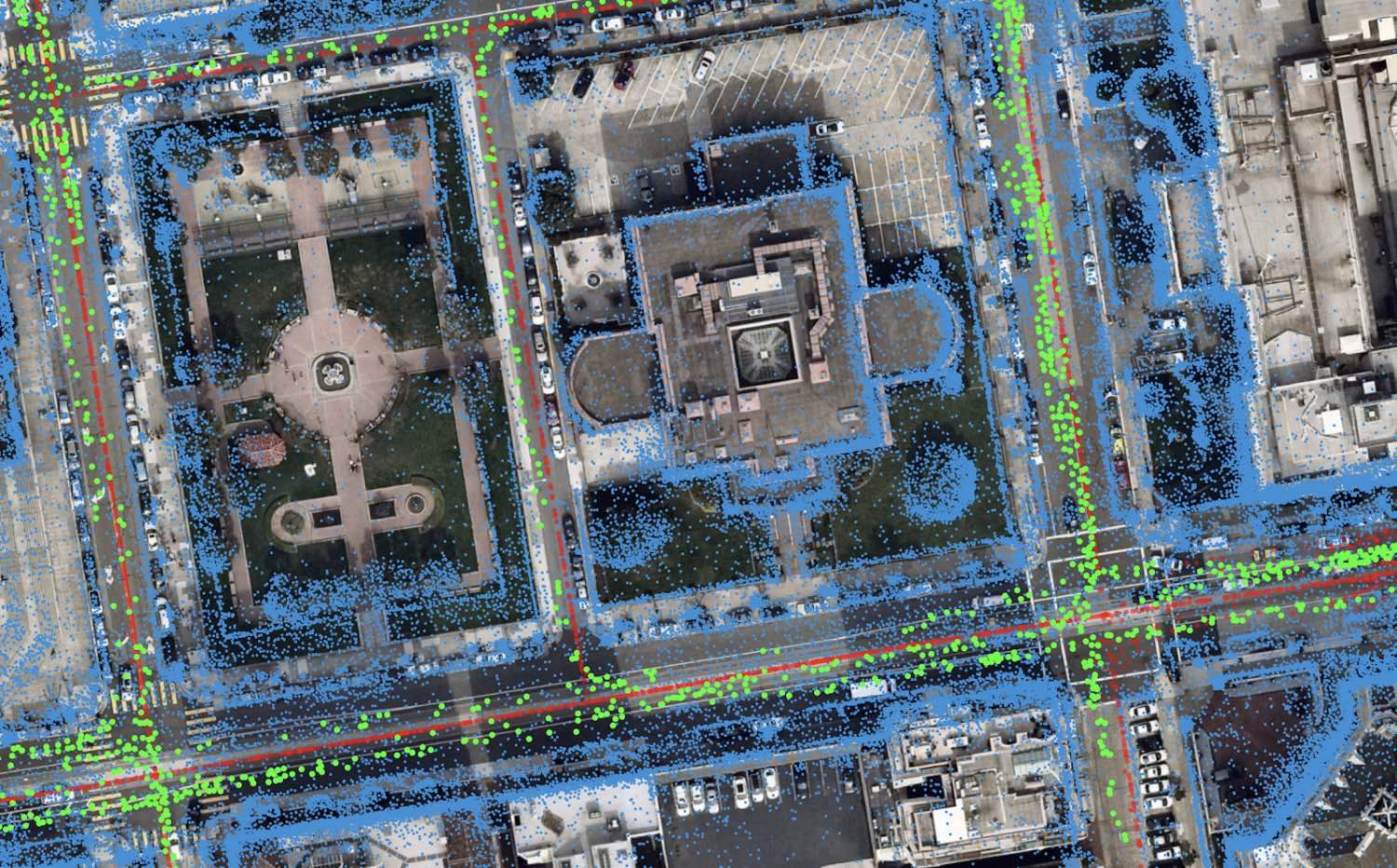}
    \caption{Detail of the geo-aligned map shown on top of satellite picture. \textcolor{blue}{Blue}: triangulated landmarks, \textcolor{green}{green}: camera poses, \textcolor{red}{red}: GPS priors.}
    \label{fig:geo-aligned}
\end{figure}

\textbf{4. Submap fusion \& map update.}
Lastly, new submaps and existing submaps in $\mathcal{M}$ are fused together rigidly. For each of $K$ submaps, we compute a 7DoF transformation $T_{k}$ that minimises the displacement between images that occur in multiple submaps (i.e., in the coordinate system of the fused map, the positions of the same image in different submaps must coincide): % VERSION BEFORE EDIT: computing individual translation offsets, rotation and scale $T_{1..K}$ of all the old and new submaps together where $K$ is the number of submaps. Unlike SfM, this is an optimisation problem over 7xK parameters $T_{1..K}$:
\begin{equation}
    \mathcal{C}(T) = \sum_i \sum_{k,l} (T_k p^k_i - T_l p^l_i) ^ 2 + \lambda \sum_{i,k} (T_k p^k_i - p_i^{gps})^2
\end{equation}
where $p_i^k$ is the pose of image $I_i$ in submap $\mathcal{M}_k$ and $k,l$ iterate over pairs of submaps containing $I_i$.

The result is a geo-aligned map (Fig.~\ref{fig:map}). Figure \ref{fig:geo-aligned} shows the map's alignment with geo-coordinates due to the combination of geometric and GPS terms.

Despite its simplicity, the overall mapping framework is remarkably robust. As further discussed in Section \ref{sec:experiments}, it robustly builds highly accurate 3D maps even if the SfM has only a 89\% success rate at the individual submap level (step 2). Moreover, both map completeness and localisation performance improve as new experiences are fused with the map, filling any low-density regions.

Removing data from the map, which might be desirable in the case of environmental change or manually detected map errors, is equally simple. The affected submaps are simply removed and we rerun the fusion steps on the remaining submaps. Note that we do not perform any global bundle-adjustment to refine positions of landmark after step 4, unlike \cite{zhu2018very}. We found that this only marginally improves map accuracy, and thus does not warrant the additional computational cost for the AR use case.

\section{Real-time SLAM synchronisation}
\label{sec:localisation}

\begin{figure}[t]
    \centering
    \includegraphics[width=80mm]{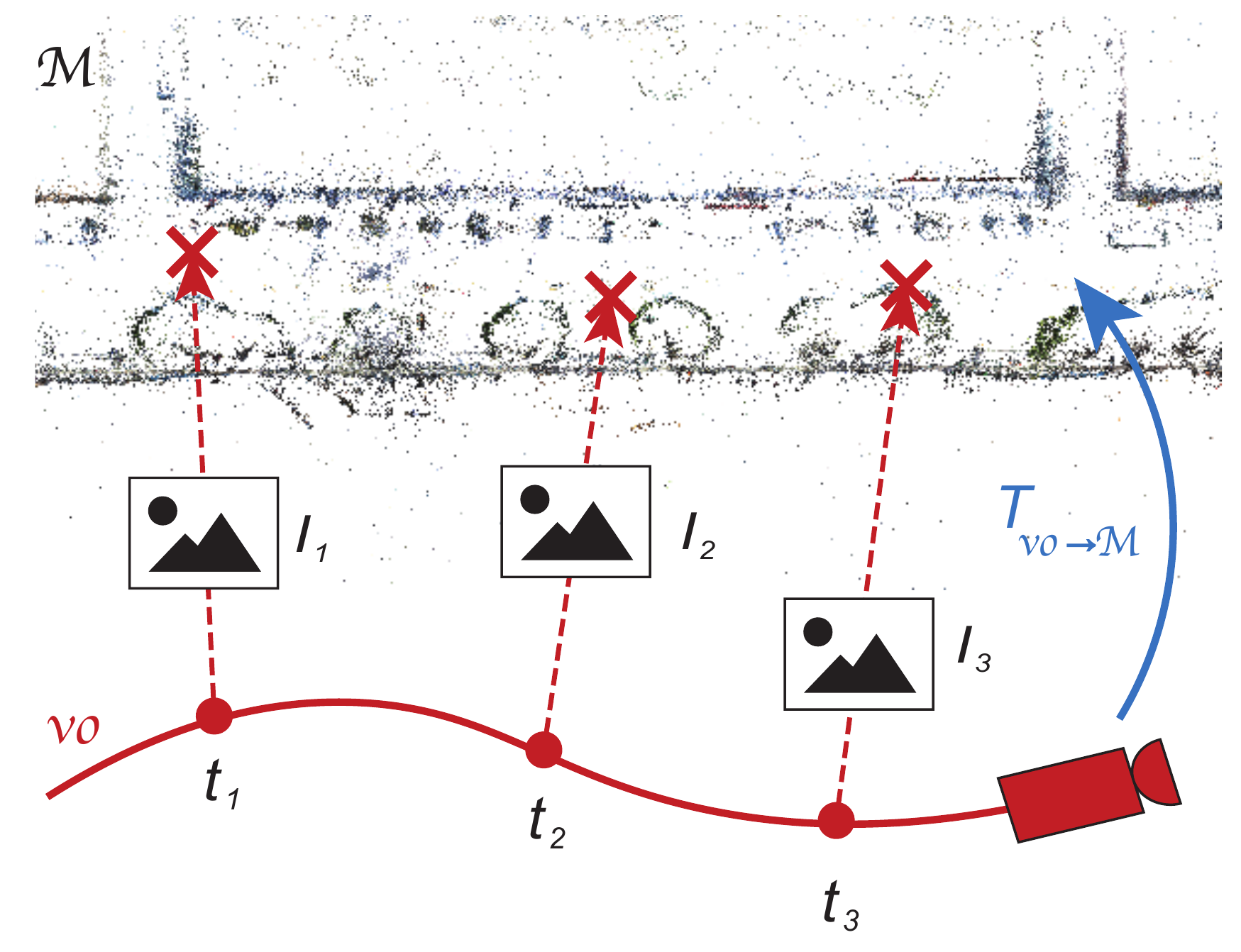}
    \caption{The process of continuously localising edge device in the map over the network. Recently captured images are transmitted over the network to determine their global position and compute optimal offset of the local visual-odometry frame of reference.}
    \label{fig:synchronisation}
\end{figure}

In this section we describe how to continuously localise an edge device in the large-scale multi-experience map we described in the previous section.

Given the map is stored in the cloud direct image-based localisation impossible. Instead, each device runs a visual-inertial odometry system that continuously tracks its own pose $p^{vo}_t$ over time in a local coordinate frame and simultaneously estimates a global-to-local coordinate transform $T_{vo \rightarrow \mathcal{M}}$. This is achieved by (1) sending recently captured keyframes over the mobile network to the cloud service containing the map, (2) the cloud service determining the edge device's global pose by performing image-based localisation, and (3) sending the global localisation result back to the edge device, as shown in Figure \ref{fig:synchronisation}. The optimal transform $T^*_{vo \rightarrow \mathcal{M}}$ is given by minimising
\begin{equation}
    \mathcal{C}(T) = \sum_{I_t} \tau^{-t}\| p_\mathcal{M}(I_t) - T p^{vo}_t \|_\delta,
\end{equation}
over the history of transmitted images $I_t$ where $p_\mathcal{M}(I_t)$ is the result of the localisation of image $I_t$ in map $\mathcal{M}$, $p^{vo}_t$ is the local pose of the image and $\tau$ is a time decay factor that favours recent measurements. We use robust Huber loss to mitigate outlier localisation. The current real-time pose is then given as
\begin{equation}
    p^\mathcal{M}_t = T^*_{vo \rightarrow \mathcal{M}} p^{vo}_t.
\end{equation}

Recall that the map is composed of submaps $\mathcal{M}_{1 \dots K}$. In order to handle a very large map, the submaps are stored together based on GPS proximity, and the image is queried against all submaps within a GPS radius. The query result with the largest number of inlier feature matches is then used. This process typically takes approximately one second per image, and is independent of the size of the map.

The accuracy of visual localisation performance depends on a number of factors and the localisation error is approximately given by
\begin{equation}
    \epsilon = \epsilon_{map} + \epsilon_{loc} + \epsilon_{vo} \frac{\Delta t}{r_{loc}}
\end{equation}
where $\epsilon_{map}, \epsilon_{loc}, \epsilon_{vo}$ is respectively the error of the map, localisation and odometry drift over localisation request time interval $\Delta t$ and $r_{loc}$ is the localisation success rate \%. As discussed in Section \ref{sec:experiments} this is usually in order of few centimeters.

\begin{figure}[bh]
    \centering
    \includegraphics[width=80mm]{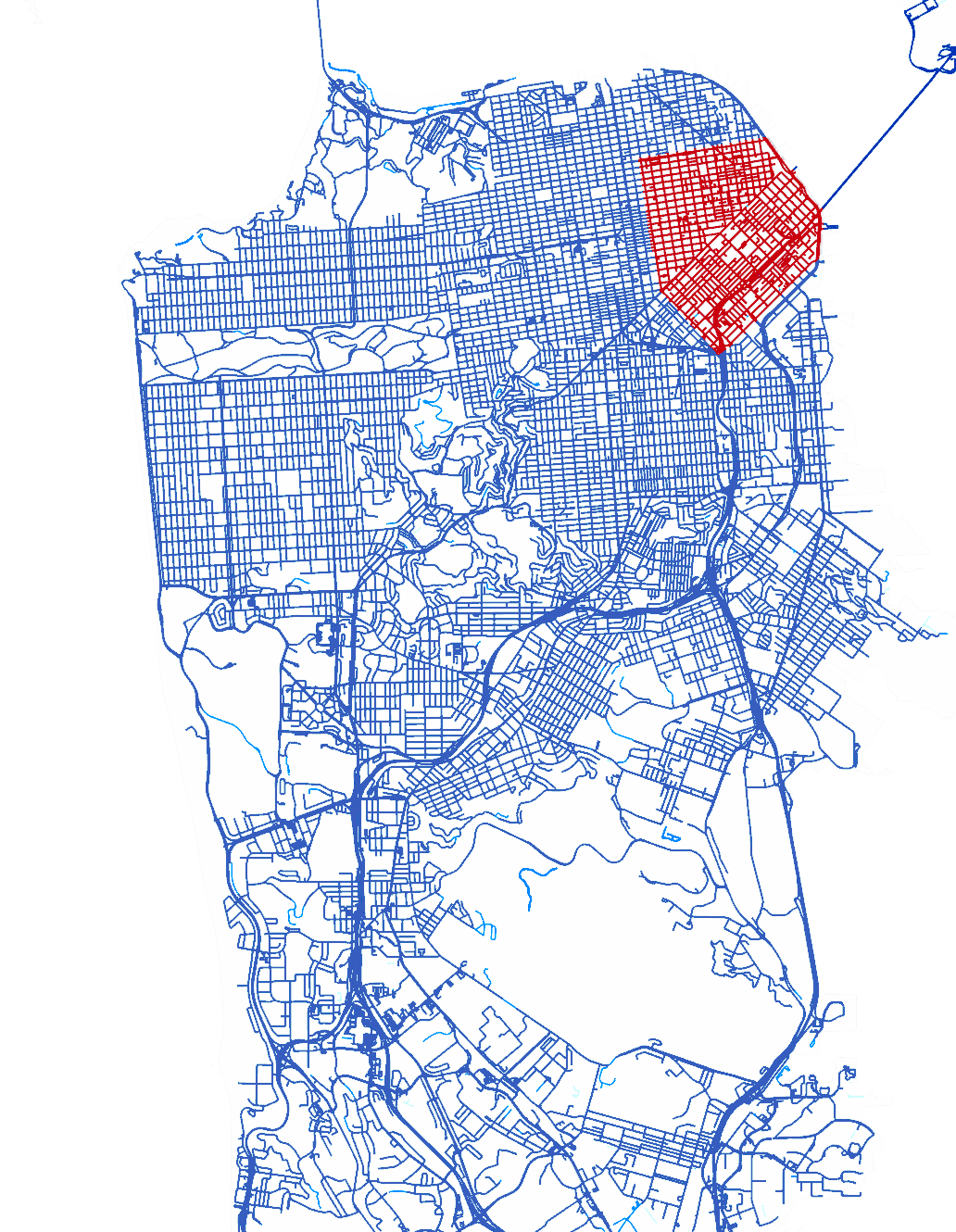}
    \caption{The dataset we collected in San Francisco and used to evaluate the system. \textcolor{blue}{Blue}: data collected by vehicles to build the map, \textcolor{red}{Red}: data collected by pedestrians to test localisation performance. The area with both vehicle and pedestrian coverage is released to the public.
    }
    \label{fig:testset}
\end{figure}

\section{Experiments}
\label{sec:experiments}

In this section we evaluate the proposed system and outline how it can be used to AR usecases. In particular, we are interested in (a) the experience-based mapping system's ability to produce large-scale maps, and (b) the edge-device localisation performance for multi-user, production AR.

\begin{table}[b]
    % \centering
    \begin{tabular}{@{}p{34mm}lll@{}}
    \hline\toprule
        \textbf{Stage} & \textbf{\# CPUs} & \textbf{Per submap} & \textbf{Total} \\
    \hline
        Data splitting & 1 & - & 1 hour\\
        Feature detection & 300 & 1 hour & 18 hours\\
        Feature matching & 300 & 1.5 hours & 27 hours\\
        SfM & 500 & 4 hours & 44 hours\\
        Submap fusion & 1 & - & 1 hour\\ \hline
        \textbf{Total} & & & \textbf{91 hours} \\
    \bottomrule\hline
    \end{tabular}
    \vspace{1mm}
    \caption{The map-building compute time broken down by each stage, for a map consisting of 7 million frames split into 4249 submaps.}
    \label{tab:buildtime}
\end{table}

\subsection{Dataset}
We have collected two datasets to build a map and evaluate the system. We report the performance of the system on the subset of the data where these two datasets overlap, and release it to the public as a part of this work.
The used dataset is several orders of magnitude larger than data used for experimentation by similar end-to-end multi-agent mapping and localisation systems (e.g., \cite{schneider2018maplab}), rendering any state-of-the-art comparisons infeasible.

\begin{figure}[t]
    \centering
    \includegraphics[width=70mm]{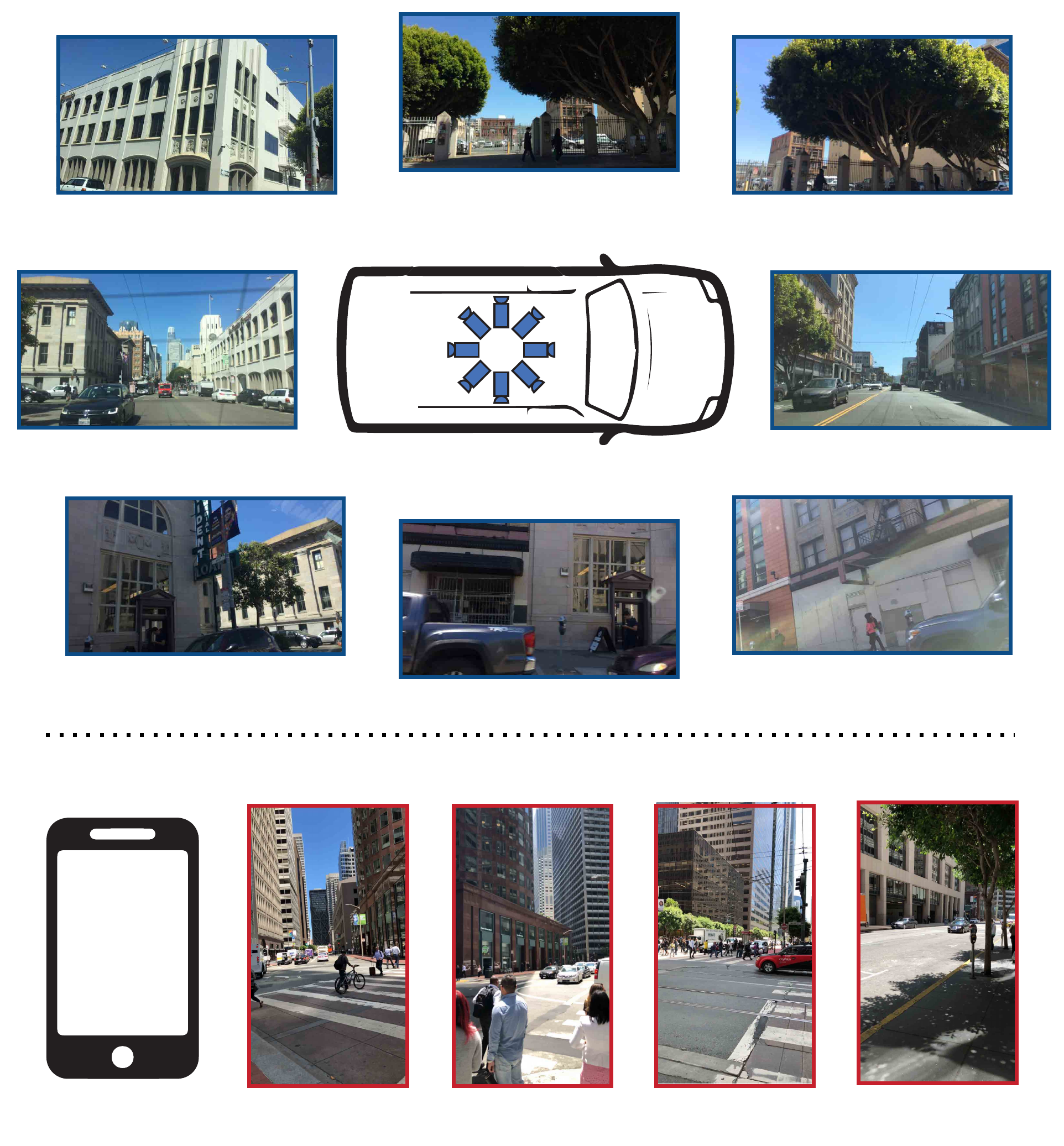}
    \caption{Samples from the dataset. \textit{Top}: a view from 8 cameras mounted on the vehicle. \textit{Bottom}: data seen by a mobile phone held by pedestrian during localisation. }
    \label{fig:data-collection}
\end{figure}

\textbf{Map-building dataset.} To build the map of the environment we collected 27M images in San Francisco using a fleet of vehicles carrying camera and GPS-equipped devices over a period of around 3 months. Subset of 7M (equivalent of roughly 1,000 miles driven) is being released and used for evaluation.

Each vehicle used for the 7M dataset was equipped by 8 rolling-shutter cameras in rosette configuration as shown in Figure \ref{fig:data-collection}. Each camera has a $70^{\circ}$ Field of View (FOV) providing a $360^{\circ}$ view around the vehicle with approximately $25^{\circ}$ overlap between neighbouring cameras. The vehicles traversed the city, covering each street multiple times and thus creating several experiences per street. Experiences come from different times of the day to cover different lighting conditions and improve the robustness of the resulting system. Most of the dataset was collected under sunny or overcast conditions. It contains images, IMU data at 100Hz, and GPS at 1Hz. The cameras are rigidly mounted but slightly different for different collect vehicles as this simplifies data collection operations. Exact camera intrinsics and extrinsics are determined as part of the structure-from-motion process and the cameras are not synchronised to further decrease the operational complexity.

\textbf{AR localisation dataset.} In order to evaluate the map’s performance for pedestrian AR use cases, we collected 1009 short video snippets of handheld data together with device's visual-inertial odometry log modelling user using the system. A human operator would stand close to the road (within 2m of the curb) and slowly looks around, to cover roughly a $180^\circ$ FOV, and then walks roughly 20m along the sidewalk. We have chosen this evaluation because for any AR experience to give a good user experience, the edge device must quickly re-localize against the map, and with minimal user interaction.

\begin{table}[t]
    %\centering
    \begin{tabular}{@{}p{17mm}p{20mm}p{25mm}p{10mm}@{}}
    \hline\toprule
        \textbf{Area} & \textbf{Attempted} & \textbf{Reconstructed} & \textbf{Yield} \\
    \hline
        1+ passes & 3000 miles & 1950 miles & 66\%\\
        4+ passes & 1517 miles & 1384 miles & 91\%\\
    \bottomrule\hline
    \end{tabular}
    \vspace{1mm}
    \caption{Map yield by number of passes. Map yield is defined as a percentage of the streets where we have successfully reconstructed batches against where we attempted to construct the map.}
    \label{tab:mapyieldpasses}
\end{table}

\begin{figure}[t]
    \centering
    \begin{tabular}{c c}
    \includegraphics[width=39mm]{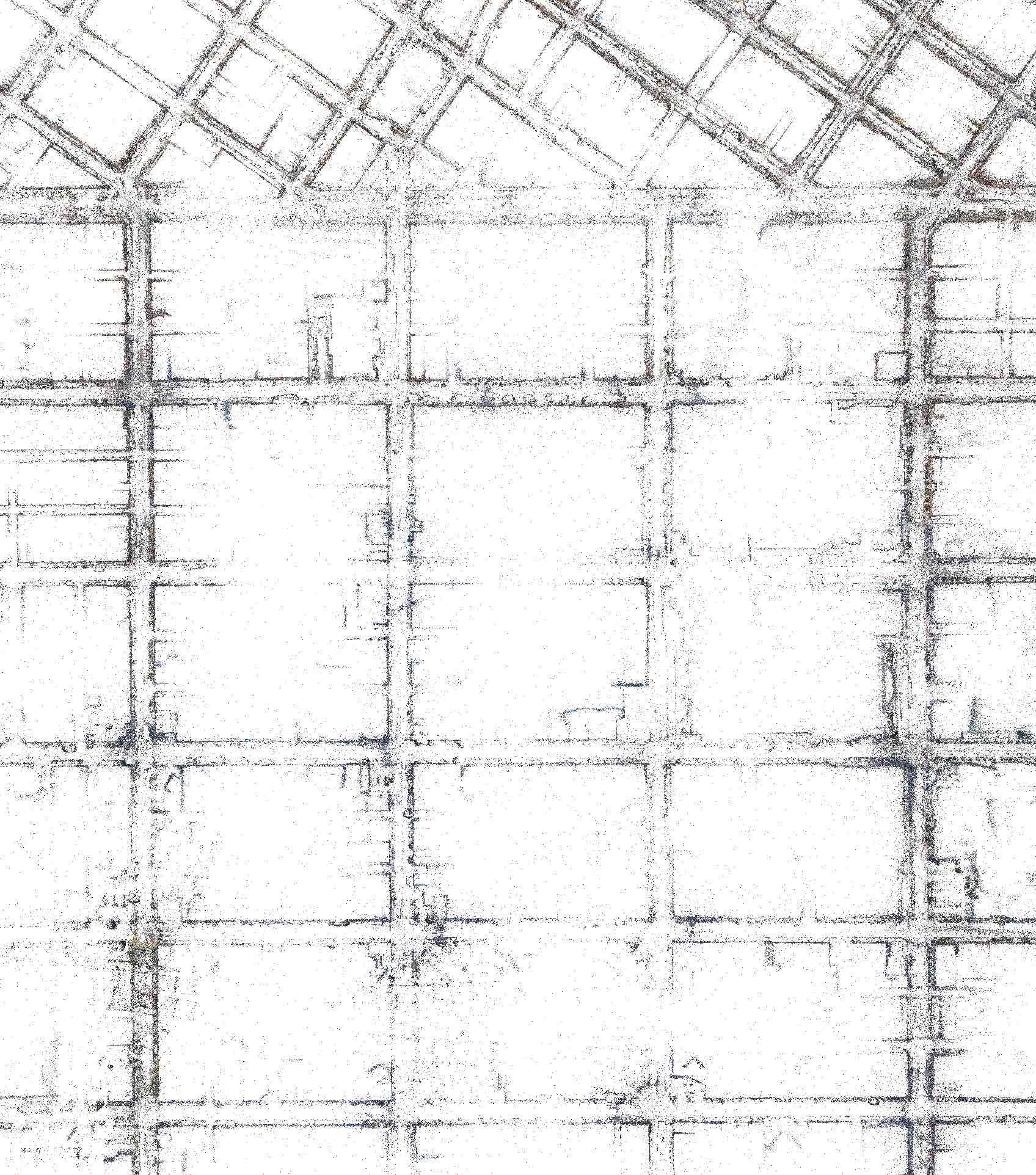} & \includegraphics[width=39mm]{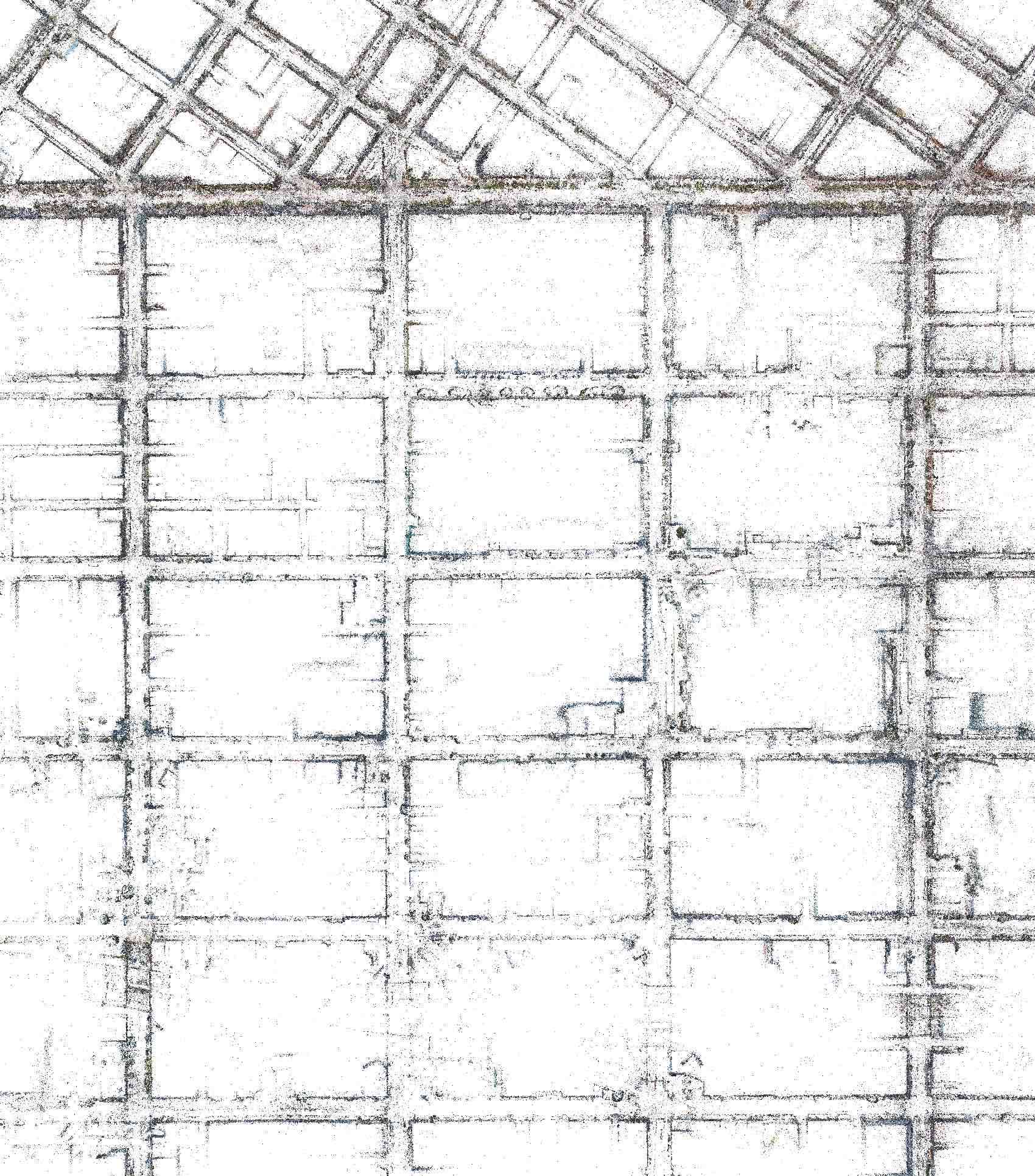}
    \end{tabular}
    \caption{Map coverage as a function of more experiences from the area.
    \textit{Left}: top-down view of a challenging area of the city after a single data collection, \textit{right}: the same area after adding more data to the map.}
    \label{fig:maplayers}
\end{figure}

\begin{figure}[b]
    \centering
    \includegraphics[width=80mm]{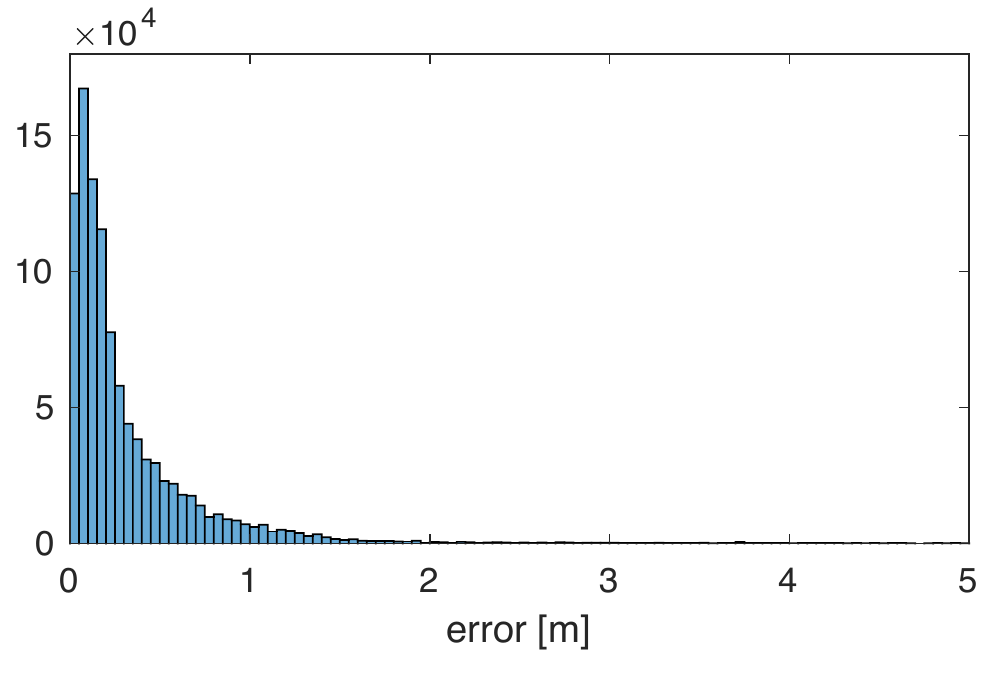}
    \caption{A histogram of metric error of localised pose, compared with a ground-truth from a high-accuracy sensor, over a 20m distance.}
    \label{fig:absolute-accuracy}
\end{figure}

\subsection{Map building performance}

%\begin{table}[b]
%    \centering
%    \begin{tabular}{|l|l|l|}
%    \hline
%        \textbf{Component} & \textbf{m} & \textbf{deg} \\
%    \hline
%        Map relative error per 10m & 0.15 & 0.20 \\
%        Visual odometry and localisation discrepancy & 0.30 & 0.61 \\
%        End-to-end relative error per 10m & 0.19 & 0.39 \\
%    \hline
%    \end{tabular}
%    \vspace{1mm}
%    \caption{Error of different stages of the system. \textcolor{red}{What is this?}}
%    \label{tab:localisationaccuracy}
%\end{table}

We processed data collected by the mapping vehicles and build the map shown in Figure \ref{fig:map}. We split the data in the evaluated area into 4249 submaps, with an average diameter of 50m, and 20m overlap between neighbouring submaps. The success rate of building submaps was 89\%. Failures were caused by difficult visual lighting conditions (strong sun), large numbers of moving objects, or operational issues with data collection hardware. We also deliberately set a high threshold for the automated and manual checks on the submaps, resulting in many being discarded.

\textbf{Implementation details.}
For our mapping pipeline, we use SIFT features \cite{LoweSIFT} and OpenMVG's incremental SfM implementation \cite{moulon2012adaptive} when building submaps. We do not perform any form of map simplification, since our submaps comfortably fit into our servers' memory, though this would be an interesting optimisation. We separately experimented with masking out dynamic agents in the images, but we do not use it for these experiment as it did not bring any improvement (parked cars can provide good features in map building and thanks to combining map localization and odometry proved enough to reliably filter outliers in localization). We do not perform further bundle adjustment for fusing the submaps as it would be infeasible given the overall map size, and this choice proved to acheive enough localisation accuracy to support AR applications (our goal). In our production system, before deploying a new map we perform a light-weight manual inspection in a web viewer to detect obvious errors (typically larger geometric disagreements between submaps in the same region which are easy to spot). After removing these submaps we rerun the submap fusion step, and deploy.

\textbf{Computation time.}
The computation took more than 35,000 hours distributed over 500 CPUs and produced a 700 GB map. For run-time and parallelisation of the different steps, see Table \ref{tab:buildtime}. Most of the time is spent in the feature matching and bundle adjustment steps, which run in quadratic and cubic time in the number of images in a submap respectively. A map update on this map consisting of building one or multiple additional submaps and fusing them in takes approximately 6 hours.

\begin{figure*}[t]
    \centering
    \includegraphics[width=58mm]{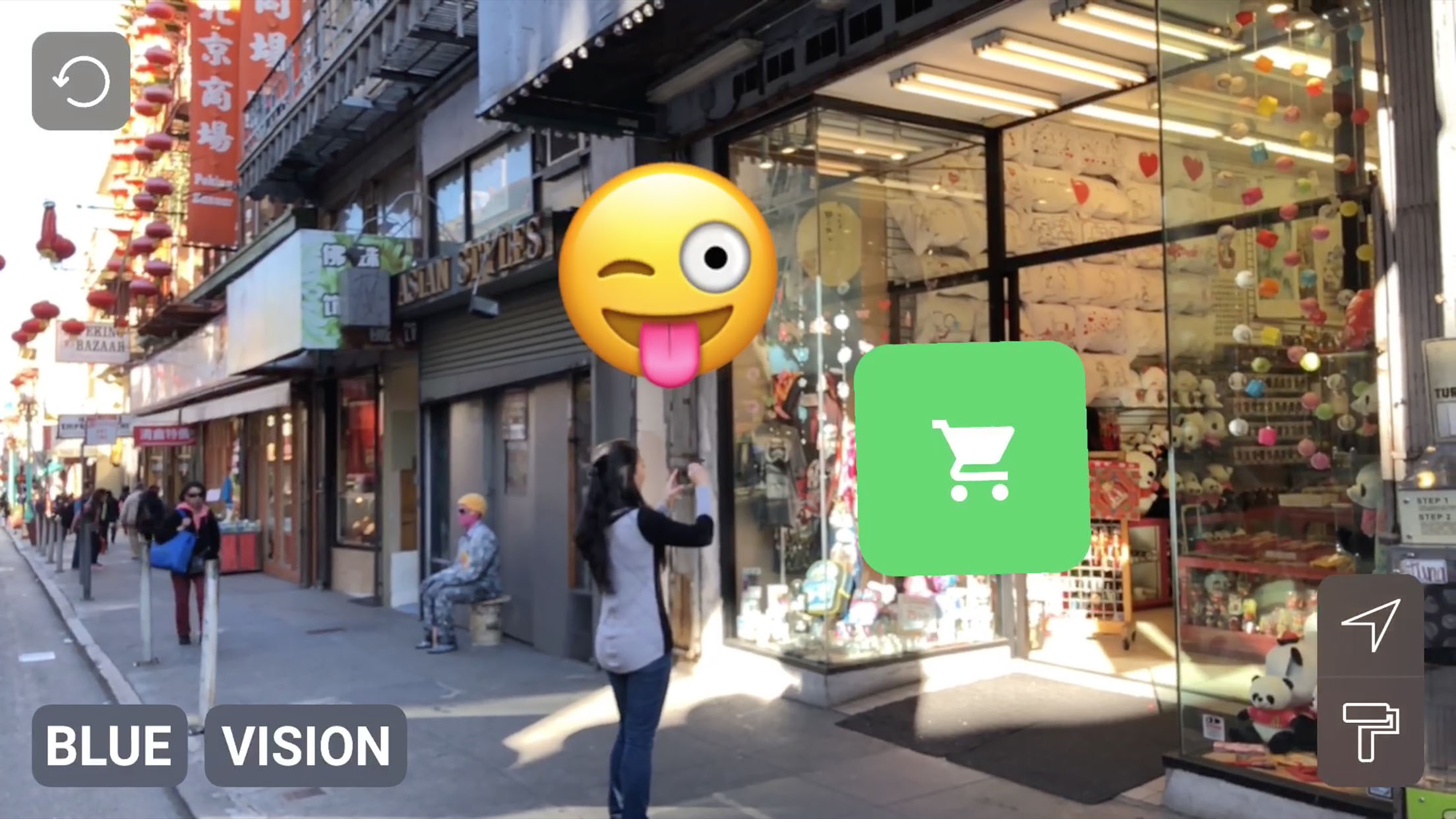}
    \includegraphics[width=58mm]{experience-gaming}
    \includegraphics[width=58mm]{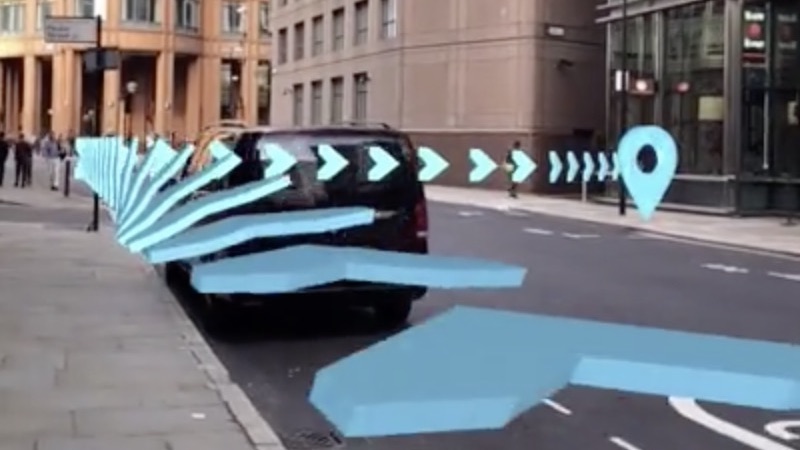}
    \includegraphics[width=58mm]{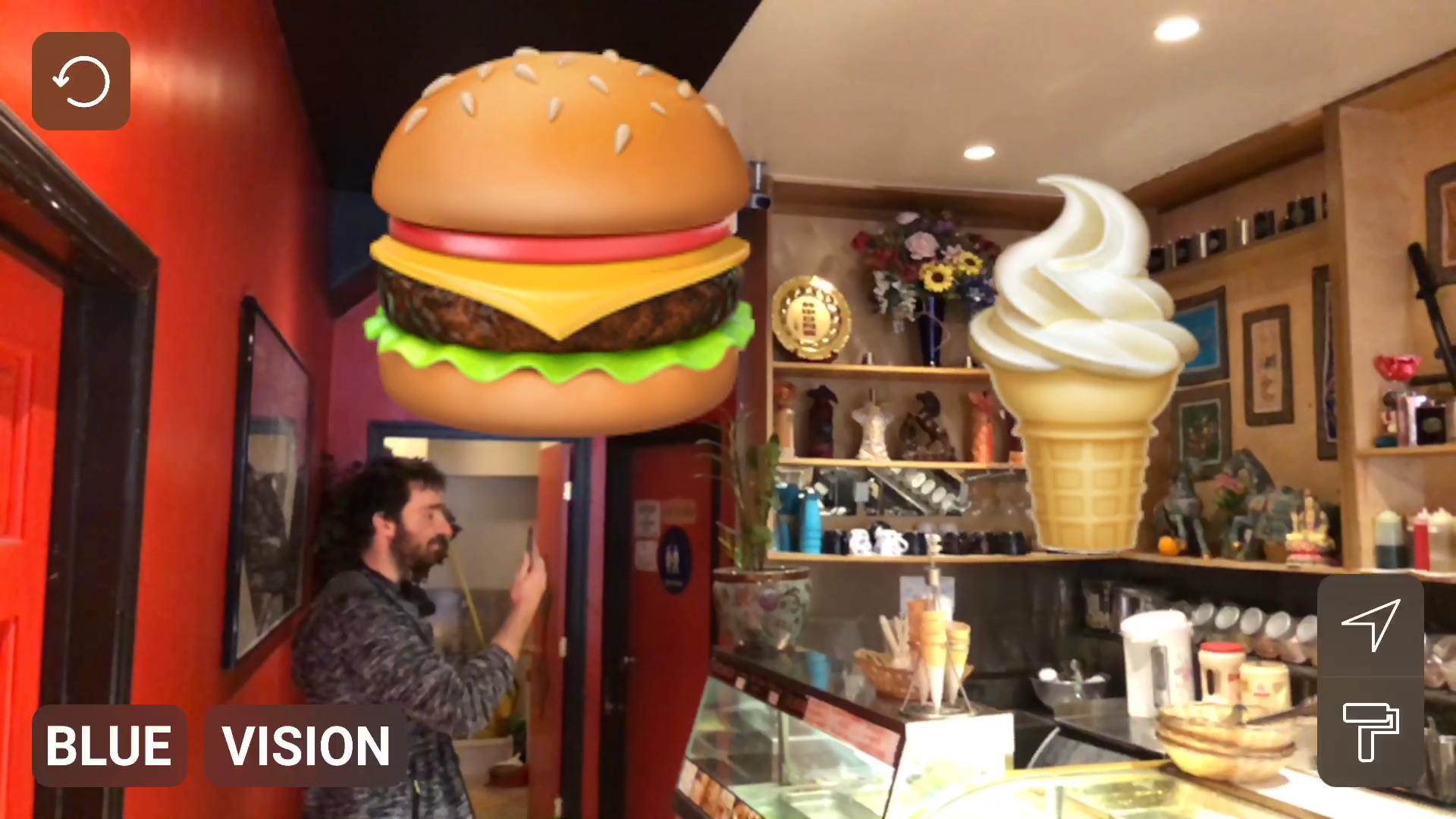}
    \includegraphics[width=58mm]{experience-social}
    \includegraphics[width=58mm]{experience-transportation}
    \includegraphics[width=58mm]{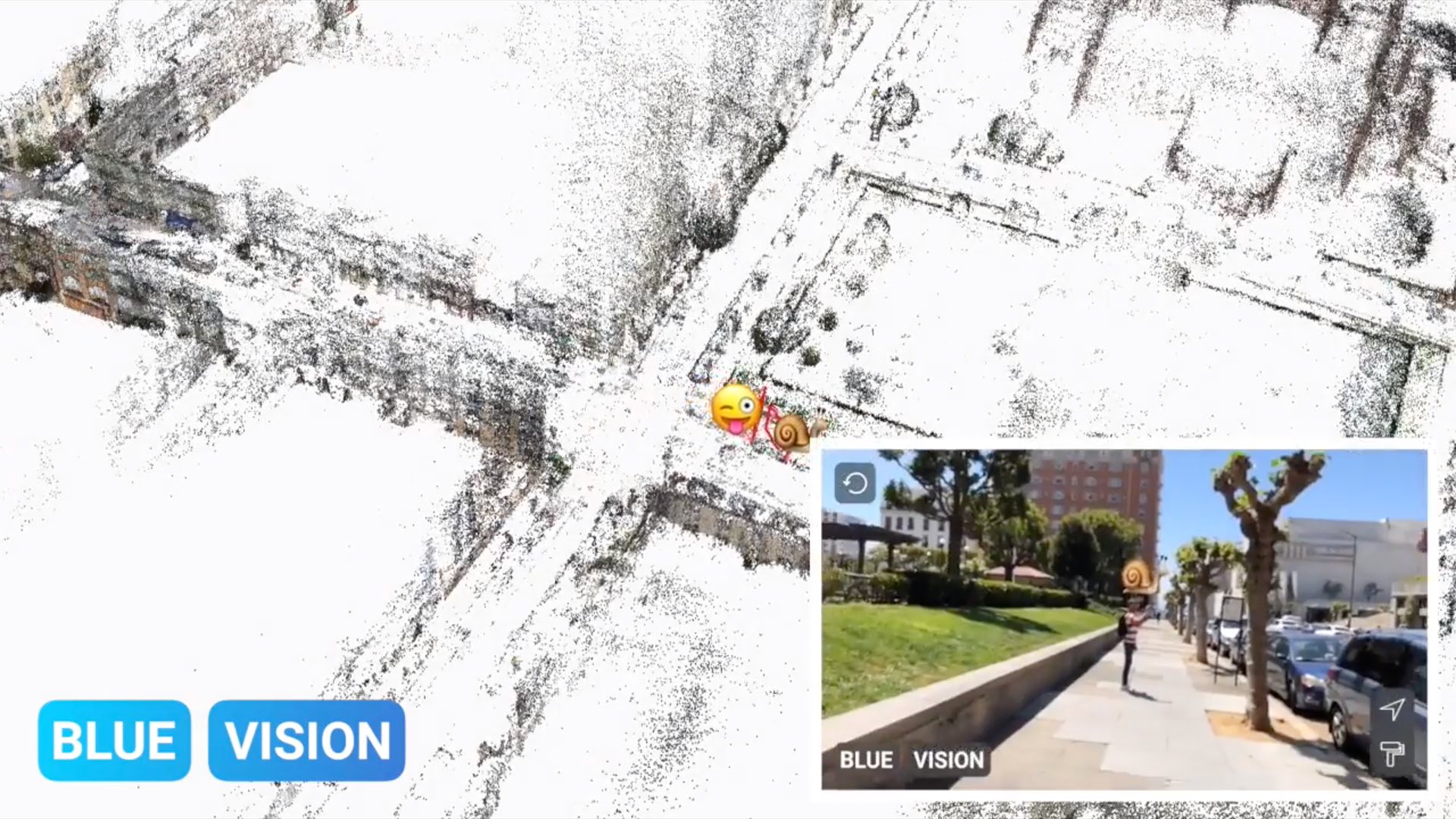}
    \includegraphics[width=58mm]{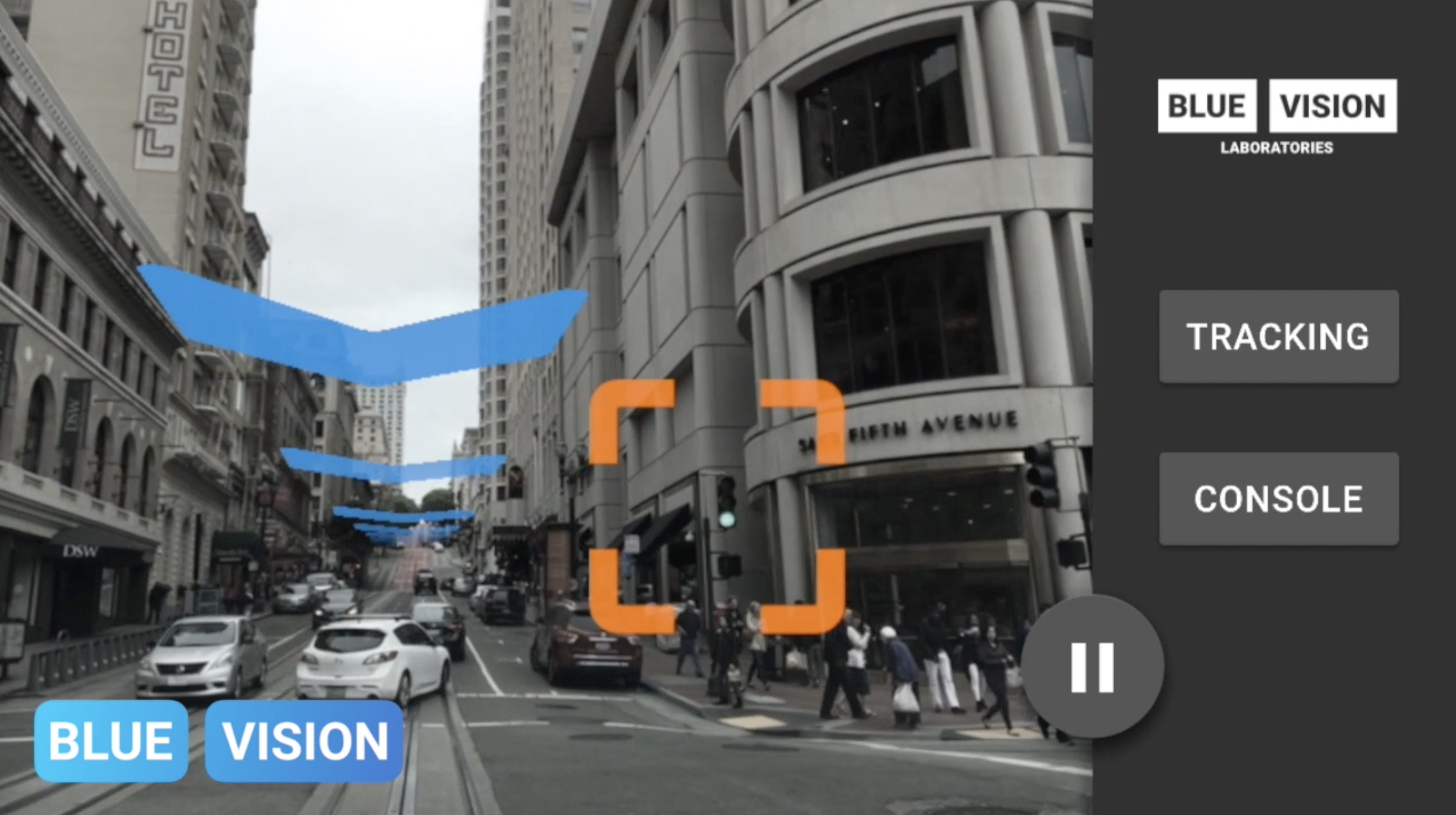}
    \includegraphics[width=58mm]{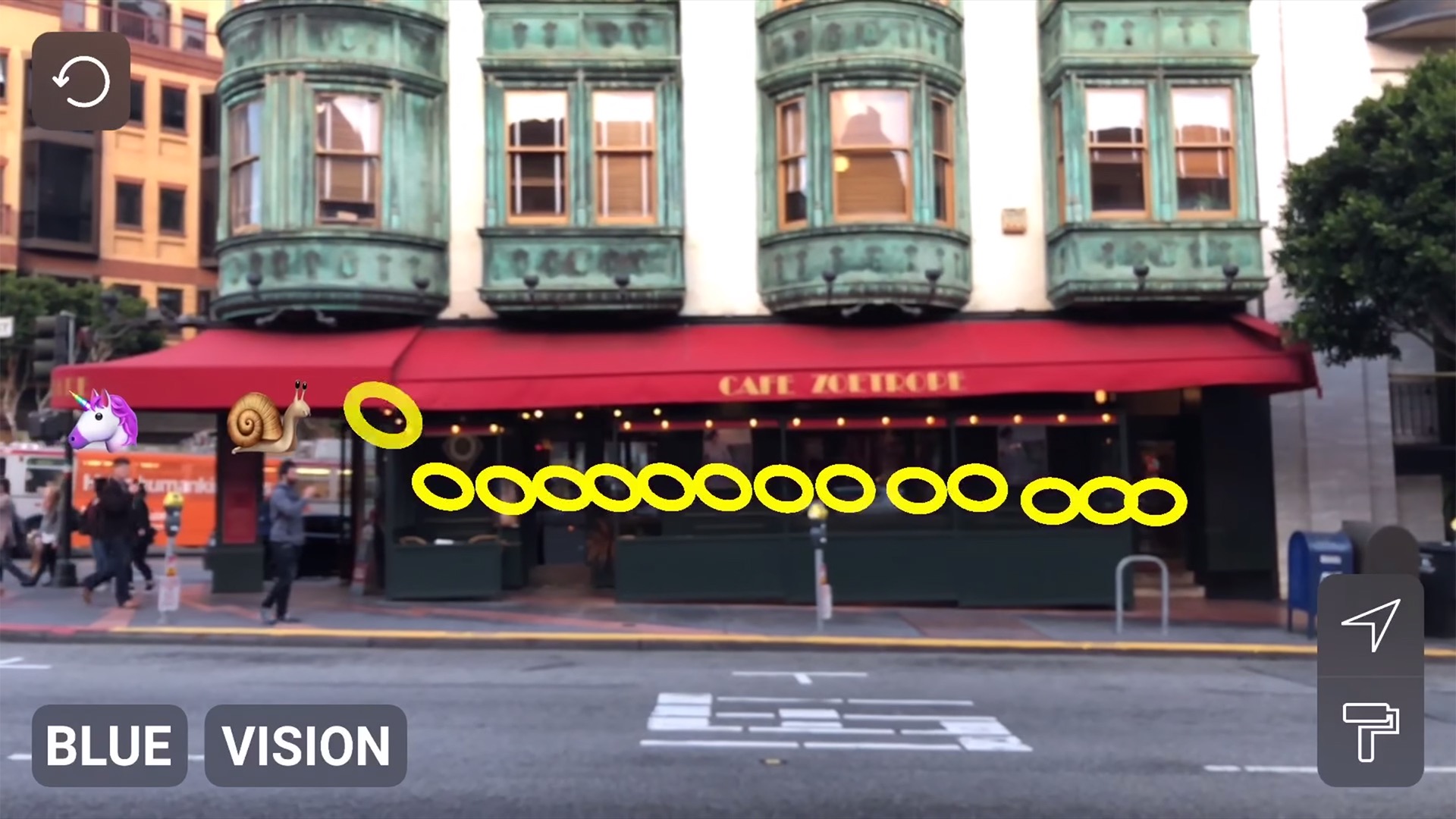}
    \caption{Examples of different augmented reality experiences we built powered by the proposed localisation of edge devices held by pedestrians or equipped on vehicles: Shared and persistent AR content visible by all users at the same location, avatar-augmented device pose, AR navigation and collaborative AR games.}
    \label{fig:ar}
\end{figure*}

\textbf{The impact of experiences on map coverage.}
Figure \ref{fig:maplayers} qualitatively shows improvement of map coverage as we add experiences. With an average of only one mapping experience per location, the map displays missing areas and would be unfit for localisation on those streets. As seen in Table \ref{tab:mapyieldpasses}, this significantly improves as more experiences are added, despite high failure rate of individual submaps.

\subsection{Localisation performance}

%\begin{figure}[b]
%    \centering
%    \includegraphics[width=80mm]{accuracy-1}
%    \includegraphics[width=80mm]{accuracy-2}
%    \caption{Translational and rotational accuracy of localisation. \lp{to find better graph}}
%    \label{fig:accuracy}
%\end{figure}

\begin{figure}[h]
    \centering
    \includegraphics[width=80mm]{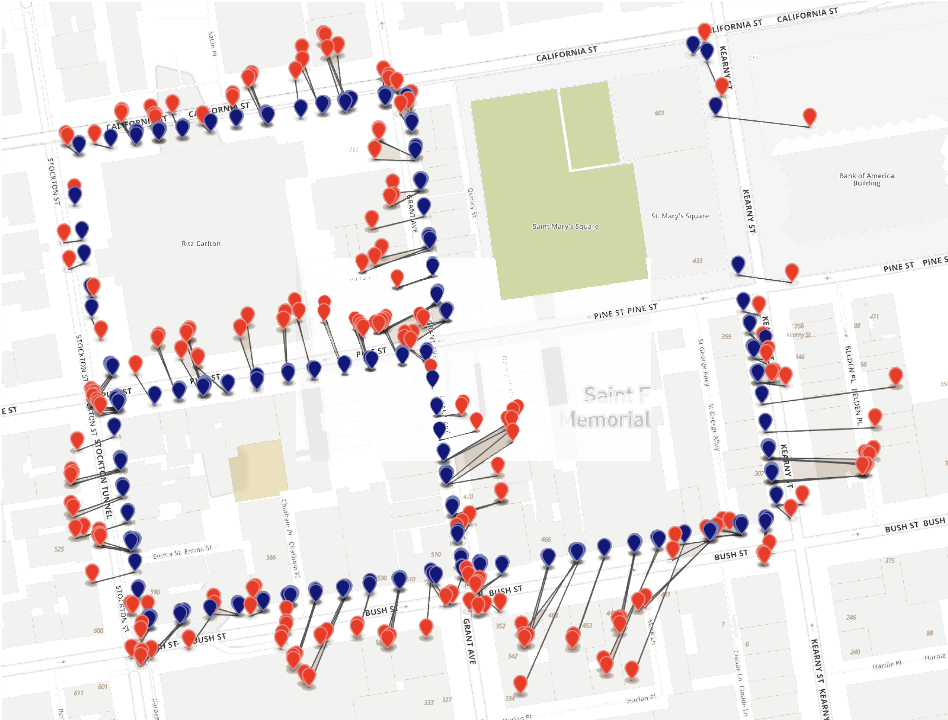}
    \caption{A comparison of typical GPS localisation accuracy of \textcolor{red}{(red)} and the proposed visual localisation \textcolor{blue}{(blue)} for a pedestrian holding a smartphone while walking down the street. Visual localisation is significantly more accurate.}
    \label{fig:gpsvsvps}
\end{figure}

We measured the performance of edge device localisation in the map. Figure \ref{fig:gpsvsvps} shows typical localisation performance for a pedestrian's smartphone as they walk down the road, as compared with GPS. The visual localisation system is significantly more accurate than GPS localisation. We used 1,009 recordings from the AR localisation dataset described above. The key localisation metric used is \emph{localisation rate}, defined as the fraction of images able to localise. We complement it with two accuracy metrics: localisation consistency and localisation accuracy.

\textbf{Implementation details.}
For the server-side localization of single images we used SIFT features \cite{LoweSIFT} and the localization tool from OpenMVG \cite{moulon2016openmvg}. We modified the approximate nearest neighbour lookup implementation to reduce the size of the data and speed of loading it into memory to fit our submap size. This was necessary to guarantee server responses within around 1 second at a stable 4G connection. For odometry, we experimented with VINS \cite{qin2017vins} and ORBSLAM \cite{murTRO2015}, but we ended up using a commercial version (ARKit \cite{arkit}), as it performed best for our handheld AR use case and provided the most robust experience on the tested smartphones.

\textbf{Localisation rate.} We evaluated 37,727 localisation requests that originated from the challenging AR dataset recordings simulating a real use of the system across the city. Out of those, 73\% were successfully localised. As these poses can be computed from erroneous matches we have also calculated that after further filtering of outlier localizations in the system, 61\% of the requests produce a coherent result when combined with the visual-inertial odometry running on the client. This rate is enough to provide a smooth user experience.

\textbf{Localisation consistency.} We measured the positional discrepancy of the server-side localisation service against the visual odometry running on the device before fusing them to be 30cm. We have tuned the system to bias heavily towards false positives in server-side localizations. The synchronisation algorithm running on the device can pick out the correct relocalisations and discard the high-variance noise. 

%\textcolor{red}{add data with "jumpiness" to show this worked better than the 90th percentile might suggest}

\textbf{Localisation accuracy.} To evaluate positional accuracy against a ground truth we collected a separate dataset comprising 2h of data. It was obtained by attaching an edge device to a self-driving-capable vehicle with high-performance odometry, and driving it around the city. We repeatedly computed the difference between raw localised poses after the device traveled distance of 20m and compared it against the groundtruth odometry. Figure \ref{fig:absolute-accuracy}, shows result of this relative error \cite{zhang2018tutorial} comparison. The mean error is 42cm (2\%) and median error is 18cm (1\%). This accuracy is sufficient for most AR usecases.

\textbf{The impact of experiences on localisation performance.} Similarly to \cite{ChurchillIJRR2013}, in Table \ref{tab:dataperformance} we observe increased localisation performance as data collected from varying times of day and weather are combined. To measure the impact on localisation performance we collected mapping data and corresponding edge device localisation recordings in one area of the map at different times of day. We evaluate the localisation rate as a function of the amount of data used for map-building.

\textbf{Computation time}
in Figure \ref{tab:testtime} summarises timing of individual stages of a single localisation request. On average the localisation requests take 970ms on 4G network and 1700ms on 3G network.

\begin{table}
    % \centering
    \begin{tabular}{@{}p{51mm}p{29mm}@{}}
    \hline\toprule
        \textbf{Mapping collections in an area} & \textbf{Localization rate} \\
    \hline
        1 day-time experience & 69.5\% \\
        4 day-time experiences  & 81.5\% \\
        6 day and night experiences & 96.6\% \\
        7 day and night experiences & 97.3\% \\
    \bottomrule\hline
    \end{tabular}
    \vspace{1mm}
    \caption{Localisation performance as a function of more experiences from the area. With more observations the localisation rate across different lighting conditions increases.}
    \label{tab:dataperformance}
\end{table}

\begin{table}
    % \centering
    \begin{tabular}{@{}p{46mm}p{34mm}@{}}
    \hline\toprule
        \textbf{Stage} & \textbf{Timing (ms)} \\
    \hline
        Image transfer & 200 (4G) -- 1,000 (3G)\\
        Feature detection & 250\\
        Feature matching & 500\\
        Triangulation & 20\\
    \bottomrule\hline
    \end{tabular}
    \vspace{1mm}
    \caption{Timing of different stages of localising one image from the cloud device in the map stored in the cloud.}
    \label{tab:testtime}
\end{table}

\subsection{AR system}
Our system can be run in parallel on multiple edge devices jointly localising in the same coordinate space of the map $\mathcal{M}$. In order to allow users to create and interact with shared persistent content and with each other, we augmented the base localisation system with two services:

\textbf{Persistent content storage} allows users to create and store the coordinates of AR content in a shared database, making it immediately visible to other users.

\textbf{Real-time pose sharing} exchanges current poses between active devices. This is important for interaction and allows you to, say, show an avatar on top of players' current locations, or play collaborative games.

Figure \ref{fig:ar} and the accompanied video showcase example AR experiences we built using these methods. Thanks to the shared coordinate system, all users see content at exactly the same place and can interact with each other.

\subsection{Limitations and failure cases}

We observed the system works robustly under usual lightning conditions. Here we summarise the most significant causes we observed to be limiting its performance:

\textbf{Multiple passes required for mapping.}
Our system achieve robustness by exploiting redundancies in the data - this was a deliberate decision. We embraced the stochastic nature of the individual methods in the overall system (each step can fail on subsets of the data), but redundancy in data allows us to handle local algorithmic failures (e.g. while building a submap from an experience with SfM can fail due poor illumination, occlusions, etc., a successful submap from another experience in the same area can fill the gap). The drawback is that we typically require a minimum amount of data in every area: note how mapping performance degrades as we go from 4 passes to 1 pass (this is the same as an experience) in Table \ref{tab:mapyieldpasses}, with localisation rate showing a similar trend in Table \ref{tab:dataperformance}. However, this is not a problem in our setup due to the scalability of our algorithms, and the simplicity of our sensor installation and data collection processes which allows us to collect large amounts of data efficiently.

\textbf{Mobile network connection.}
One of the biggest limitations we observed in practice is the requirement of a stable mobile data connection during its use. We observed common sudden drops in mobile bandwidth when testing in crowded areas of the city. This has a large noticeable impact on initalisation of the system and also on the positional error accumulated due to dead-reckoning drift of odometry. This can be solved by doing the map relocalisation on the edge device. While possible, it would require optimising the algorithm's runtime, using quicker (most likely binary rather than float) feature descriptors, and simplification of the map to reduce its size to serve it to the device, which will also need to cache parts of the map. This optimisation is left as future work for the next versions of the system.

\textbf{Viewpoint change too large.}
While the system is resilient to localization from some novel viewpoints, the second most common issue comes from inability of the system to match images due to occlusion, or when there is a large difference in viewpoint. The latter is mostly noticeable in areas where there is a larger gap between the sidewalk (where the edge device is held) and the street (where the mapping collections happens). Thanks to the submap implementation and focus on data redundancy, one way to tackle this is to also use data collected from the sidewalks with a smaller hand-held camera rig for mapping (additionally to our current car mapping collect). This works well, however, pedestrian data collection can become a bottleneck when scaling to larger areas. An alternative is to explore methods like ASIFT \cite{asift} known to increase the robustness to viewpoint changes.

\begin{figure}
    \centering
    \includegraphics[width=41mm]{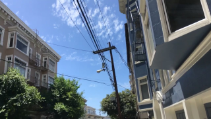}
    \includegraphics[width=41mm]{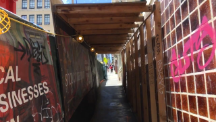}
    \includegraphics[width=41mm]{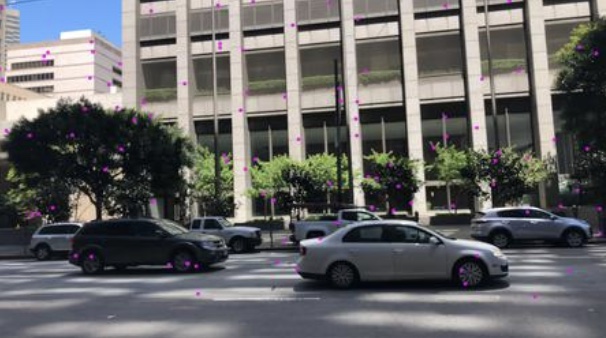}
    \includegraphics[width=41mm]{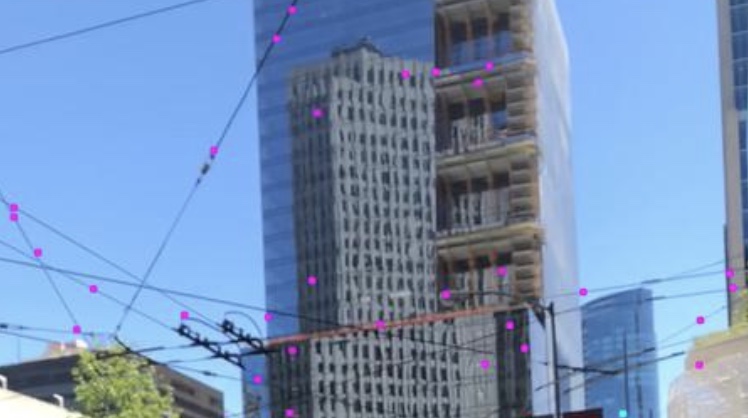}
    \caption{Typical localisation failure cases. These include obstructed sidewalks, user pointing the phone upwards, repetitive structures and reflections.}
    \label{fig:failures}
\end{figure}

\textbf{Other failure cases.}
Figure \ref{fig:failures} displays some of the failure cases mentioned above, such as obstructed sidewalks, and a few more failures due to excessive vertical tilting, reflections, and repetitive textures. These can prevent or slow down an initial localisation in some areas (e.g. where recent changes happened due to constructions, or the user is pointing their phone towards the sky), but we observed that their impact on the overall experience is mitigated by the VIO system.

\section{Conclusions}
We have presented the first published end-to-end production system for building collaborative augmented reality experiences at city-scale. The presented system can be extended in many ways, for example, by automatically updating the map when detecting changes in the environment, potentially with data transmitted by the edge devices. We believe that , together with the released dataset, this work will enable the community to address these and other challenges, and accelerate the development of AR systems towards seamless and ubiquitous experiences.

\acknowledgments{
This work was done thanks to the hard work and support of the Blue Vision Labs team: 
Ben Haines,
Bryon Shannon,
David Evans,
Ed Dingley,
Fady Kalo,
Filippo Brizzi,
Gabriel Nica,
Gabriele Angeletti,
George Thomas,
Giacomo Dabisias,
Guido Zuidhof,
Haoyue Zhu,
Ivan Katanic,
James Close,
Jesse DaSilva,
Karim Shaban,
Long Chen,
Lorenzo Peppoloni,
Lucas Chatham,
Matej Hamas,
Mimosa Nguyen,
Miranda Aperghis,
Owen Parker,
Robert Kesten,
Sina Nickdel,
Stepan Simsa,
Suraj Mannakunnel Surendran,
Yerzhan Utkelbayev.
}

\bibliographystyle{abbrv-doi}

\bibliography{main}

\end{document}